\documentclass{article} % For LaTeX2e
\usepackage{colm2024_conference}
\usepackage{xspace}

\usepackage{microtype}
\usepackage{hyperref}
\usepackage{url}
\usepackage{booktabs}
\definecolor{darkblue}{rgb}{0, 0, 0.5}
\hypersetup{colorlinks=true, citecolor=darkblue, linkcolor=darkblue, urlcolor=darkblue}

\usepackage{algorithm}
\usepackage{algpseudocode}
\usepackage{amsmath}
\usepackage{times}
\usepackage{latexsym}
\usepackage{graphics}
\usepackage{graphicx}

\usepackage{tabularx,colortbl}
\usepackage{booktabs}
\usepackage{tabularx}
\usepackage{makecell}
\usepackage{multirow}
\usepackage{float}
\usepackage{listings}

\usepackage{booktabs}
\usepackage{pifont} % For checkmarks and crosses
\usepackage{soul}
\usepackage{listings}
\usepackage{xcolor}
\usepackage{arydshln} % For dashed lines
\usepackage{wrapfig}

\lstdefinelanguage{PythonLike}{
    morekeywords={for, in, if, else, elif, while, def, return, import, as, from, class, self, True, False, None},
    sensitive=true,
    morecomment=[l]{\#},
    morestring=[b]",
}

\newcommand{\cmark}{\ding{51}}  % Checkmark symbol
\newcommand{\xmark}{\ding{55}}  % Cross symbol

\newcommand{\mehdi}[1]{{\color{black}#1}}
\newcommand{\mingyu}[1]{{\color{black}#1}}

\title{X-EcoMLA: Upcycling Pre-Trained Attention into MLA \\ for Efficient and Extreme KV Compression}

\author{Guihong Li$^*$, Mehdi Rezagholizadeh$^*$, Mingyu Yang\thanks{ Equal Contribution First Authors, with order determined alphabetically. } , Vikram Appia, Emad Barsoum \\
Advanced Micro Devices, Inc. (AMD)\\
\texttt{\{guihong.li,mehdi.rezagholizadeh,mingyu.yang\}@amd.com} \\
}
\newcommand{\ours}{{X-EcoMLA}\xspace}

\begin{document}

\maketitle

\begin{abstract}
Multi-head latent attention (MLA) is designed to optimize KV cache memory through low-rank key-value joint compression. Rather than caching keys and values separately, MLA stores their compressed latent representations, reducing memory overhead while maintaining the performance. While MLA improves memory efficiency without compromising language model accuracy, its major limitation lies in its integration during the pre-training phase, requiring models to be trained from scratch. This raises a key question: \textit{can we use MLA’s benefits fully or partially in models that have already been pre-trained with different attention mechanisms?} 
In this paper, we propose \ours~to deploy post training distillation to enable the upcycling of Transformer-based attention into an efficient hybrid MLA variant through lightweight post-training adaptation, bypassing the need for extensive pre-training.
We demonstrate that leveraging the dark knowledge of a well-trained model can enhance training accuracy and enable extreme KV cache compression in MLA without compromising model performance. 
The experimental results show that our proposed method can effectively compress the KV cache while preserving the performance on the benchmarks; specifically, for Llama3.2-1B-Instruct baseline, a 6.4× compression achieves the same average score by using only 3.6B training tokens and 70 GPU hours on AMD MI300, whereas a 10.6× compression have less than 0.1\% average score drop with 7B training tokens and 140 GPU hours. The code for this work is available at \url{https://github.com/AMD-AGI/AMD-Hybrid-Models}.
\end{abstract}

\section{Introduction}

Large language models (LLMs) have become ubiquitous, revolutionizing both academic research and industrial applications~\cite{brown2020language,openai2023gpt4,chowdhery2022palm,liu2024deepseek,guo2025deepseek}. Their success largely stems from pre-training and instruction tuning on vast amounts of data, the power of self-attention in Transformer architectures, and the computational capabilities of GPU accelerators. 
% However, this success comes with challenges, particularly the high memory overhead  and the computational complexity of Transformer models, which lead to increased deployment costs and may limit their accessibility and broader application.

Despite their widespread adoption, Transformers face two major challenges in their self-attention mechanism: quadratic computational complexity and high memory demands for KV cache storage, particularly when processing long sequences.
Significant efforts have been made to mitigate these challenges by: (a) replacing self-attention with new sub-quadratic architectures such as state-space models (SSMs)~\cite{gu_mamba_nodate, dao2024transformers,poli2024mechanistic}; (b) enhancing the efficiency of existing Transformer self-attention mechanisms~\cite{arora2024simple,ainslie2023gqa,zhang2024the,yang2024gated, qin2024transnormerllmfasterbetterlarge}; and (c) developing hybrid solutions that combine the advantages of both quadratic and sub-quadratic models~\cite{lieber2024jamba,dong2024hymbahybridheadarchitecturesmall,wang2024mamba,bick2024transformers}. Among these approaches, the second category is particularly appealing, as it requires minimal architectural changes {while capitalizing on existing hardware optimized for Transformers}. This work focuses on improving the efficiency of existing self-attention mechanisms within this category.

% \begin{wrapfigure}[25]{r}{0.5\textwidth}
\begin{figure}
  \centering
   \includegraphics[width=0.95\columnwidth]{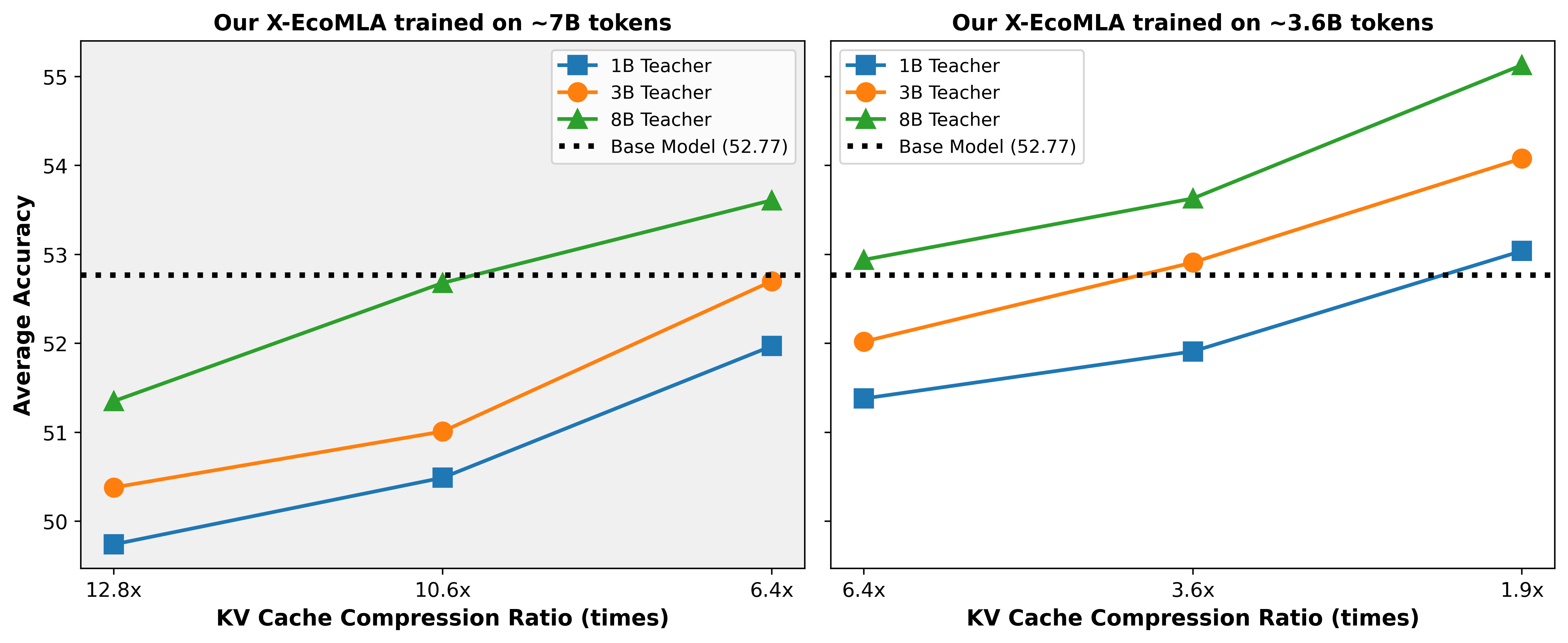}
   \vspace{-0.05in}
  \caption{ \small{ \ours with Different Teacher Sizes: [Right] Our results show that using Llama3.2-1B, 3B, and 8B teacher models enables KV cache compression of Llama3.2-1B by 1.9×, 3.6×, and 6.4× using 3.6B tokens respectively, without compromising average accuracy across multiple tasks on the LM Harness Evaluation benchmark. {[Left] With ~7B training tokens, we can further compress the KV cache to 10.6× and 12.8x, while maintaining competitive accuracy. }}}
  \label{fig:first plot}
  \vspace{-15pt}
\end{figure}

% \end{wrapfigure}

Multi-head attention (MHA) is a fundamental mechanism in Transformer architectures. However, during inference, MHA requires {the saving of} large {amount of} key-value (KV) cache, resulting in high memory consumption. To address this {challenge}, DeepSeek~\citep{liu2024deepseek, liu2024deepseek-v3} recently proposed multi-head latent attention (MLA), a novel approach that compresses the KV cache while maintaining the performance of LLMs.
The mainstream training paradigm for MLA has relied on pre-training from scratch using vast amounts of data and computational resources. {For example, pre-training the Deepseek-v3 model required 2.664M GPU hours on Nvidia H800 clusters.}
% \footnote{For example, training the Llama 3.2-1B model required 370K GPU hours on Nvidia H100 80GB devices \href{https://huggingface.co/meta-Llama/Llama-3.2-1B}{HuggingFace Link}}. 
This highlights a major challenge: developing models with a new attention mechanism demands significant computational resources during pre-training. Given the substantial effort already invested in training Transformer models, a natural question arises—can we transfer the rich pre-training knowledge {from trained LLMs} into more efficient MLA models without training them from scratch?

% There is evidence in the literature confirming the feasibility of knowledge transfer {with} architectural adaptations {for LLMs}, including MambaInLlama~\citep{wang2024mamba}, MOHAWK~\citep{bick2024transformers}, and HedgeHog~\citep{zhang2024the}.  
{Existing approaches showcase strong evidence supporting the feasibility of knowledge transfer with architectural adaptations for LLMs, such as MambaInLlama~\citep{wang2024mamba}, MOHAWK~\citep{bick2024transformers}, and HedgeHog~\citep{zhang2024the}.}
Inspired by these solutions, {we introduce \ours, a cost-effective knowledge transfer approach designed to upcycle pre-trained multi-head attention into MLA.}
% Inspired by these solutions, we propose \ours, an economical knowledge transfer approach for upcycling pre-trained multi-head attention into MLA. 
In \ours, we initialize MLA from its corresponding pre-trained attention {blocks} using our static or dynamic SVD approach, followed by distillation from a well-trained teacher model.
{By leveraging the dark knowledge of a high-quality model, we enhance training accuracy and achieve extreme KV cache compression in MLA without sacrificing performance.  Our results demonstrate that an 8B teacher model enables 6.4×  KV cache compression of the Llama3.2-1B-Inst baseline while preserving 100\% of its average score across multiple tasks on the LM Harness Evaluation benchmark. This requires only 3.4B training tokens and about only 70 GPU hours on AMD MI300, while pre-training the Llama3.2-1B model requires 370K GPU hours\footnote{https://huggingface.co/meta-Llama/Llama-3.2-1B}.  Furthermore, we achieve a 10.6× compression with 7B training tokens and around 140 GPU hours, with less than 0.1\% average score drop.}
We summarize our {major} contributions as follows:
\begin{itemize}
\item We propose a lightweight post-training approach to upcycle pre-trained attention to MLA, significantly reducing computational costs by eliminating the need for training from scratch.
\item {We develop static and dynamic SVD-based initialization techniques to improve the convergence and accuracy of MLA adaptation.}
\item {We demonstrate that leveraging a larger teacher model enables extreme KV cache compression while maintaining model performance, achieving up to 10.6× compression with minimal accuracy loss.} (see Fig.~\ref{fig:first plot}).
\item {We validate the effectiveness of our approach through extensive experiments on the LM Harness Evaluation benchmark, showcasing its efficiency across various settings and LLMs.}
\end{itemize}

\vspace{-7pt}

\section{Related Work}
\vspace{-5pt}
{ This section gives a brief overview of related work; a detailed version is available in Appendix~\ref{related_work}.
\vspace{-7pt}
\paragraph{KV Cache Management in Transformers} Transformers store key-value (KV) vectors for each token and each layer during auto-regressive generation, leading to high memory requirements during inference, especially for long sequences. Several methods have been proposed to reduce the KV cache size in Transformers, broadly divided into training-based and post-training approaches~\cite{shi2024keep}. Post-training techniques are easier to apply but may degrade performance due to information loss. These include KV eviction (e.g., Heavy Hitter~\cite{zhang2023h2o}), sliding window attention~\cite{arora2024simple, beltagy2020longformer}, \mehdi{low-rank projection approach for KV-cache compression~\cite{chang2025palu},} and quantization~\cite{kang2024gear, zhang2024pqcache}. Some strategies, like Attention Sink~\cite{xiao2023efficient} and KV merging~\cite{wang2024model}, aim to mitigate information loss while keeping memory usage low. This paper, however, focuses on training-based solutions, which tend to offer a better trade-off between efficiency and accuracy.
\vspace{-7pt}
\paragraph{Training-based KV Cache Management} Training-based methods modify attention mechanisms to reduce memory use during inference. Multi-query attention (MQA)\cite{shazeer2019fast} and grouped-query attention (GQA)\cite{ainslie2023gqa} reduce KV cache size by sharing keys/values among heads. YOCO~\cite{sun2024you} reduces redundancy with a shared KV cache across layers. DeepSeek-V2~\cite{liu2024deepseek} introduces multi-head latent attention (MLA), which compresses hidden states via low-rank projection, reducing cache size while outperforming standard MHA. Inspired by MLA’s efficiency, this work explores adapting MLA to pre-trained model and we try to address this question: Can we upcycle pre-trained models to their MLA counterparts without costly retraining?
\vspace{-7pt}
\paragraph{Upcycling Attention} Upcycling refers to upgrading pre-trained models with minimal computation~\cite{komatsuzaki2022sparse}. In attention upcycling, existing attention blocks are adapted into efficient forms like MLA without full retraining. GQA~\cite{ainslie2023gqa} and Hedgehog~\cite{zhang2024the} achieve this via light fine-tuning or distillation. Hybrid models like MambaInLlama~\cite{wang2024mamba} and MOHAWK~\cite{bick2024transformers} distill knowledge from Transformer attention into Mamba layers. 
\mehdi{MHA2MLA~\cite{ji2025towards} introduces a data-efficient fine-tuning approach for converting MHA to MLA via partial RoPE removal and joint SVD-based low-rank approximation. In contrast, \ours adopts a unified RoPE design with a shared Key-RoPE vector across all heads (similar to DeepSeek MLA), and employs structured initialization along with knowledge-distillation-based efficient training to enable effective MLA-based upcycling.}
}

\vspace{-5pt}

\section{Background}

In this section,  we formalize the mathematical framework of MHA and MLA, following the notation from the original DeepSeek-V2 technical report~\cite{liu2024deepseek} with some slight modifications.

\subsection{Multi-Head Attention (MHA)}
MHA projects the input hidden state $H$ into three distinct spaces using learned weight matrices:
\begin{equation}
    Q = H W^Q, \quad K = H W^K, \quad V = H W^V, 
\end{equation}
where  $ H \in \mathbb{R}^{l \times d} $ is the input sequence representation with $l$ being sequence length and $d$ internal hidden state dimension, and $ W^Q, W^K, W^V \in \mathbb{R}^{d \times n_h d_h} $ are the learnable projection matrices where $n_h$ is the number of attention heads, and $d_h$ is the head dimension. The attention scores and final outputs are computed as:
\begin{equation}
    \begin{split}
        & A = \text{Softmax} \left( \frac{Q K^T}{\sqrt{d}} \right), ~~~ O = A V W^O
    \end{split}
    \label{eq:attention}
\end{equation}
where $W^O \in \mathbb{R}^{d \times d}$ is the output transformation matrix. During inference, MHA requires caching $ K $ and $ V$ for all past tokens, leading to a storage requirement of $2 n_h d_h l$.  

\subsection{Multi-Head Latent Attention (MLA)}\label{subsec:mla}
%\mehdi{In this section, at eq. 4, we introduce $d_{qk}$ but in the previous subsection, we use the notation of $d_h$ for the head dimension for all three Q, K and V. }

MLA introduces a low-rank joint compression strategy for keys and values, reducing the KV cache size.
Instead of caching $K$ and $V$, MLA compresses them into a lower-dimensional latent representation $C^{KV}$:
\begin{equation}
    C^{KV} = H W^{DKV},
\end{equation}
where $ W^{DKV} \in \mathbb{R}^{d \times r_{kv}} $ is the down-projection matrix, and $ r_{kv} \ll d_h n_h$ is the compressed dimension for keys and values. The keys and values are then reconstructed from $ C^{KV}$ using:
\begin{equation}
    K^C = C^{KV} W^{UK}, \quad V^C = C^{KV} W^{UV}, 
\end{equation}
where $ W^{UK} \in \mathbb{R}^{r_{kv} \times n_h d_{qk}}$ and $W^{UV} \in \mathbb{R}^{r_{kv} \times n_h d_h} $ are the up-projection matrix for keys and values. Notice that in this paper we consider a more flexible setting where the queries and keys could have different dimensionality $d_{qk}$ other than $d_h$. During inference, the learned matrices can be absorbed into the existing projection layers: $W^{UK} $ and $W^{UV}$ can be merged into $W^Q $ and  $W^O$ respectively. 

In the meantime, such low-rank compression can be also applied to the queries to reduce the memory usage while training:
\begin{equation}
    C^{Q} = H W^{DQ}, \quad Q^C = C^{Q}W^{UQ},
\end{equation}
where $C^{Q} \in \mathbb{R}^{l \times r_q}$ represents the compressed latent vector for queries, $r_q$ denotes and query compression dimension, $ W^{DQ} \in \mathbb{R}^{d\times r_q}$ is the down-projection matrix and $W^{UQ} \in \mathbb{R}^{r_q \times n_h d_{qk}}$ denotes the up-projection matrix.

However, such low-rank KV compression is not compatible with Rotary Position Embedding (RoPE) as it breaks the matrix absorbing mechanism. As a result, the authors in ~\cite{liu2024deepseek} propose decoupled Rotary Position Embeddning where additional multi-head queries $Q^R$ and a shared key $K^R$ are applied to carry RoPE, which can be expressed as:
\begin{equation}
    Q^R = \text{RoPE}(C^Q W^{QR}), \quad K^R = \text{RoPE}(H W^{KR}),
\end{equation}
where $W^{QR} \in \mathbb{R}^{r_q \times n_h d_r}$ and $W^{KR} \in \mathbb{R}^{r_q \times d_r}$ represent the matrices to produce the decoupled keys and queries. Then, the RoPE embeddings $Q^R, K^R$ and Non-RoPE (NoPE) embeddings $Q^C, K^C$ are concatenated to perform the attention operation:
\begin{equation}
Q = [Q^C; Q^R], \quad K = [K^C; \text{repeat}(K^R)],
\end{equation}
where $\text{repeat}(.)$ denotes duplicating $K^R$ for each head. After concatenation, the same attention operation is applied as in Equation \ref{eq:attention}. During inference, MLA requires caching only $ C^{KV} $ and $K^R$, reducing the storage requirement to  $(r_{kv}+d_r) l$ which is significantly smaller than the standard MHA cache size.

\section{Methodology of \ours }

We begin with a pre-trained Transformer model referred to as the "base model". Our methodology in this paper concerns upcycling the attention modules in the base model into MLA modules to save the KV cache memory while remaining minimum training efforts and performance degradation. To achieve that, we first propose our SVD-based weight initialization approach to better inherit the knowledge from the pre-trained model. Additionally, our initialization approach offers both static and dynamic rank selection. After initialization, we adopt the knowledge distillation training process in MambaInLlama~\cite{wang2024mamba}, which includes: 
end-to-end knowledge distillation, and direct preference optimization (DPO).

\subsection{SVD-based Weight Initialization}
As introduced in Section \ref{subsec:mla}, the MLA module comprises several key parameters: {$W^{DQ}$, $W^{UQ}$, $W^{QR}$, $W^{DKV}$, $W^{KR}$, $W^{UK}$, and $W^{UV}$.} Correct initialization of these matrices is crucial to ensure a smooth transition and to preserve as much of the original model's knowledge as possible. For other parameters such as the output transformation matrix $W^O$ and the feedforward module, we directly copy them from the pre-trained base model as initialization.

%the down-projection matrix for queries $W^{DQ}$, the up-projection matrix for NoPE queries $W^{UQ}$, the up-projection matrix for RoPE queries $W^{QR}$, the down-projection matrix for keys and values $W^{DKV}$, the down-projection matrix for RoPE keys $W^{KR}$, the up-projection matrix for NoPE keys $W^{UK}$, and the up-projection matrix for values $W^{UV}$. Correct initialization of these matrices is crucial to ensure a smooth transition and to preserve as much of the original model's knowledge as possible. For other parameters such as the output transformation matrix $W^O$ and the feedforward module, we directly copy them from the pre-trained base model as initialization.

Although MLA and MHA are fundamentally different, MLA closely approximates a low-rank version of MHA if we disregard the positional embedding and intermediate layernorms.  
Based on this observation, we propose {an} SVD-based initialization method that initializes the MLA weights using SVD-decomposed low-rank matrices derived from the pre-trained attention weights. Our method is simple and can be illustrated within few lines of pseudocode as shown in Algorithm \ref{alg:pseudocode}. As demonstrated in our experiments, such straightforward initialization could significantly enhance the knowledge distillation performance compared to random initialization.

For simplicity, we assume the pre-trained attention block employs MHA with weight matrices $W^Q, W^K, W^V \in \mathbb{R}^{d \times d_h n_h}$. However, our method can be readily extended to MQA \cite{shazeer2019fast} and GQA \cite{ainslie2023gqa} by duplicating the key and value weight matrices to align with the total number of attention heads. Our approach supports varying query and key {dimensions} under the constraint $d_{qk} = d_{r} \leq d_h$. For the value {dimension}, we assume that the MLA and MHA modules share the same dimensionality $d_h$. 

To initialize $W^{DQ}$, $W^{UQ}$ and $W^{QR}$, we begin by performing SVD on the pretrained weight matrix $W^Q$ to obtain:
\begin{equation}
    W^Q = U_q \Sigma_q V^T_q, 
\end{equation}
where $U_q \in \mathbb{R}^{d \times r_q}$, $\Sigma_q \in \mathbb{R}^{r_q \times r_q}$, and $V_q \in \mathbb{R}^{d_h n_h \times r_q}$. For the down-projection matrix $W^{DQ}$, we directly set $W^{DQ} = U_q$. For the up-projection matrices, we first compute $\Sigma_q V^T_q$ and reshape it to $\overline{W}^{UQR} \in \mathbb{R}^{r_q \times n_h \times d_h}$. Then we partition it along the last dimension to derive the two up-projection matrices: $W^{UQ}$ for the first $d_{qk}$ elements and $W^{QR}$ for the last $d_{r}$. This can be expressed as: 
\begin{equation}
    W^{UQ} = \text{reshape}(\overline{W}^{UQR}[:,:,:d_{qk}]), \;\;
    W^{QR} = \text{reshape}(\overline{W}^{UQR}[:,:,-d_r:]),
\end{equation}
where $\text{reshape}(.)$ denotes the reshape function that merges the last two dimensions of the input tensor.

For the remaining MLA weight matrices associated with keys and values, we first perform a joint SVD on the concatenated $W^K$ and $W^V$:
\begin{equation}
    [W^K, W^V] = U_{kv} \Sigma_{kv} V^T_{kv}, 
\end{equation}
where $U_{kv} \in \mathbb{R}^{d \times r_{kv}}$, $\Sigma_{kv} \in \mathbb{R}^{r_{kv} \times r_{kv}}$, and $V_{kv} \in \mathbb{R}^{2 d_h n_h \times r_{kv}}$. For the down-projection matrix $W^{DKV}$, we directly set $W^{DKV} = U_{kv}$. To derive the up-projection matrix $W^{UV}$, we set it equal to the last $d_h n_h$ columns of $\Sigma_{kv} V'_{kv}$. For the up-projection matrix for keys $W^{UK}$, we first extract the first $d_h n_h$ columns of $\Sigma_{kv} V'_{kv}$ and reshape them into $\overline{W}^{UK} \in \mathbb{R}^{r_{kv} \times n_h \times d_h}$. Subsequently, we select the first $d_{qk}$ elements along the last dimension of $\overline{W}^{UK}$ and reshape it back to obtain $W^{UK}$, which can be expressed as:
\begin{equation}
    W^{UK} = \text{reshape}(\overline{W}^{UK}[:,:,:d_{qk}]).
\end{equation}

Lastly, for the RoPE key embedding matrix $W^{KR}$, we employ a different initialization strategy as all attention heads share the same RoPE key embedding in MLA. We first compute the average key projection matrix $W^K_{avg} \in \mathbb{R}^{d \times d_h}$ across all attention heads. Then, we extract the last $d_r$ columns from it to initialize $W^{KR}$, which can be expressed as: 
\begin{equation}
    W^{KR}=W^K_{avg}[:, -d_{r}:].
\end{equation}

Clearly, it is crucial to determine the rank values \( r_q \) and \( r_{kv} \) for the $Q$ and $KV$ matrices. We propose two methods for the rank selection:  
(i) \textbf{Fixed Rank Selection}, where constant rank values are used across all transformer blocks.  
(ii) \textbf{Dynamic Rank Selection}, which determines rank values based on two predefined energy thresholds {\( \delta_q, \delta_{kv}  \in (0,1] \)}.  Taking \( W_Q \) as an example, we use SVD to decompose it and obtain the singular values:  
\begin{equation}
    \sigma_j, \quad j \in \left[ \min(d, n_h d_h) \right], \quad \text{where } \sigma_j \geq \sigma_{j+1}
\end{equation} 
To determine the optimal rank, we define the cumulative energy based on the squared singular values. The rank \( r_{q_i} \) for the \( i \)-th transformer block is selected as follows:  
\begin{equation}
r_{q_i} = \arg\min_R \left\{ \sum_{j=1}^{R} \sigma_j^2 \geq \delta_q E \right\}, \quad \text{where } E = \sum_{j=1}^{\min(d,\ n_h d_h)} \sigma_j^2
\end{equation}

This approach ensures that the selected rank captures at least {\( \delta_q \)} fraction of the total energy \( E \). The same energy-based dynamic rank selection process can be applied to the KV weight matrix as well.

% \begin{algorithm}[t]
% \begin{minted}{python}
% # MHA weights: W_Q, W_K, W_V
% # MLA weights: W_DQ, W_UQ, W_QR, W_DKV, W_UK, W_KR, W_UV

% # Initialization of W_DQ, W_UQ, and W_QR
% U_q, sigma_q, V_q = svd(W_Q)
% W_DQ = U_q
% W_UQR_bar = (sigma_q @ V_q).view(r_q, n_h, d_h)
% W_UQ = W_UQR_bar[:, :, :d_qk].view(r_q, n_h*d_qk)
% W_QR = W_UQR_bar[:, :, -d_r:].view(r_q, n_h*d_r)

% # Initialization of W_DKV, W_UK, W_KR, W_UV
% U_kv, sigma_kv, V_kv = svd(torch.cat((W_K, W_V), -1))
% W_DKV = U_kv
% W_K_avg = W_K.view(d, n_h, d_h).mean(1)
% W_KR = W_K_avg[:, -d_r:]

% W_UKV = sigma_kv @ V_kv
% W_UK_bar = W_UKV[:, :d_h*n_h].view(r_kv, n_h, d_h)
% W_UK = W_UK_bar[:,:,:d_qk].view(r_kv, n_h*d_qk)
% W_UV = W_UKV[:, d_h*n_h:]
% \end{minted}
% \caption{ Python-like pseudocode of the proposed SVD initialization for MLA}
% \label{alg:pseudocode}
% \end{algorithm}

\subsection{Training Process}

\paragraph{End-to-End Knowledge Distillation}  The primary training stage involves an end-to-end distillation using a Supervised Fine-Tuning (SFT) dataset introduced in~\cite{wang2024mamba}. During this stage, the goal is to minimize the KL divergence loss between the outputs of the student model (i.e., \ours) and the teacher model, which can be expressed as:
\begin{equation}
    \mathcal{L}_{\theta} = \sum_{t=1}^T \mathbf{KL}[p(\cdot | y_{1:t}, x, \theta_T) || p(\cdot|y_{1:t}, x, \theta) ], 
\end{equation}
where $\theta$ is the trainable parameters of the student model and $\theta_T$ is the parameters of the teacher model, which is kept frozen while the distillation process. Such distillation step is crucial for transferring the rich, pre-trained knowledge from the teacher model. In Section \ref{sec:ce}, we show that this distillation process is more effective than plain cross-entropy loss. Note that the teacher model can be any strong, pre-trained model—even one that is different from the base Transformer model that is used to construct the student model.
\vspace{-5pt}
\paragraph{Direct Preference Optimization (DPO)} In the final training stage, we perform DPO which is a binary cross entropy loss to adjust the preference of the student model. Following ~\cite{wang2024mamba}, we set the distilled student model itself as the reference model as it makes the training much stabler and produces a sufficient performance gain, which can be observed from the experiments section.     
\vspace{-5pt}

\section{Experiments}
\vspace{-5pt}
\subsection{Experimental Setup}

\paragraph{Model} In this paper, we {primarily focus on SmolLM-series models \footnote{https://huggingface.co/collections/HuggingFaceTB/smollm-6695016cad7167254ce15966} (SmolLM-135M-Instruct, SmolLM-360M-Instruct, SmolLM-1.7B-Instruct) and Llama 3-series models \footnote{https://huggingface.co/collections/meta-Llama/Llama-32-66f448ffc8c32f949b04c8cf} (Llama3.2-1B-Instruct, Llama3.2-3B-Instruct, Llama3.1-8B-Instruct) as our base models which span a variety of scales.  All the models employ GQA as their attention module. } 
For MLA, we utilize the same number of attention heads and the same head dimension for values. We tried multiple $r_{q}$, $r_{kv}$, and $d_{qk} (d_{r})$ while {maintaining} a similar amount of parameters as the base model.

% primarily focus on the Llama3.2-1B-Instruct model~\cite{grattafiori2024Llama_31} as our base model, which contains $16$ layers with GQA modules. For each GQA module, there exists $32$ attention heads, $8$ attention heads for keys and values, and a head dimension of $64$. For MLA, we utilize the same number of heads and the same head dimension for values. We tried multiple $r_{q}$, $r_{kv}$, and $d_{qk} (d_{r})$ while remaining a similar amount of parameters as the base model.
As mentioned earlier, we employ knowledge distillation, transferring knowledge from teacher models to the base model. In our experiments, we explore the impact of different teacher model sizes, including Llama3.2-1B-Instruct, Llama3.2-3B-Instruct, and Llama3.1-8B-Instruct~\cite{grattafiori2024Llama_31}. This approach allows us to analyze how the varying sizes of the teacher model impact the performance and efficiency of the distilled student model. %(self-distillation), Llama3.2-3B-Instruct, and Llama3.1-8B-Instruct~\cite{grattafiori2024Llama_31}. This approach allows us to analyze how the varying sizes of the teacher model impact the performance and efficiency of the distilled student model.
\vspace{-5pt}
\paragraph{Training procedure} 
Our approach begins with a pre-trained model, so we focus primarily on post-training rather than pre-training the base model again. We use a two-stage training procedure. Specifically, in the first stage (knowledge distillation), our \ours model is trained with {SFT} using the KL loss between the output of the base model and the teacher model. We use AdamW optimizer with $\beta=(0.9, 0.98)$ and set the training batch size at 96. We use the same dataset as used in the paper~\cite{wang2024mamba}. It is constructed by aggregating and reformatting data from multiple publicly available sources, including \textit{OpenHermes-2.5}~\cite{OpenHermes2_5}, \textit{GenQA}~\cite{chen2024genqa}, and \textit{Infinity-Instruct}~\cite{infinity_struct}. For the second stage (DPO), based on finetuned models, we use the combination of three datasets for DPO preference tuning: \textit{Llama3-ultrafeedback}~\cite{ultrafeedback_armorm}, \textit{orca\_dpo\_pairs}~\cite{OpenOrca}, and \textit{ultrafeedback\_binarized}~\cite{cui2023ultrafeedback}. We set the training batch size at 64 and still use AdamW as the optimizer. {We unfreeze all parameters in our \ours model and train them for one epoch for both SFT and DPO.} %We unfreeze all parameters in the base model and train them for one epoch for each of our proposed models.

\begin{table*}[t]
     \setlength\extrarowheight{2pt}
    \centering
    \scalebox{0.64}{
    \begin{tabular}{llccccccccccc}
        \toprule
        Model and Setting & Init. Method & KV-Size & ARC & ARE & HS & MMLU & OBQA & PIQA & PBMD & RA & WG & Avg. \\        \midrule
        % \multicolumn{13}{c}{100\% MLA Layers- Teacher: Identical to the base Model} \\        
        \midrule
        SmolLM135M-Inst & Base Model & 100\% & 27.39 & 43.64 & 41.91 &	24.11 &	33.80 &	67.30 &	55.80 &	32.06 &	51.22 & \textbf{41.91} \\ 
        \hdashline
        $\uparrow$\ours & Random ($r_{kv}=160$)  & 50\% & 21.08 & 30.68 & 25.27 & 22.95 & 25.00 & 53.54 & 55.20 & 22.87 & 49.57 & 34.03 \\
        $\uparrow$\ours + DPO & Random ($r_{kv}=160$)  & 50\% &  21.59 & 30.56 & 25.17 & 22.95 & 26.20 & 53.97 & 55.20 & 22.97 & 49.33 & 34.22 \\
        $\uparrow$\ours & Fixed ($r_{kv}=160$) & 50\% &   27.05 &	43.31 &	40.10 &	24.51 &	31.80 &	66.97 &	55.40 &	31.39 &	51.30 &	41.31 \\ 
        $\uparrow$\ours + DPO & Fixed ($r_{kv}=160$) & 50\% &  28.24 & 45.33 & 40.26 & 26.20 & 33.60 & 65.89 & 55.60 & 32.63 & 50.36 & \textbf{42.01} \\ 
        $\uparrow$\ours  & Dynamic ($\delta_{kv}=0.85$) & 49.5\% &  26.88 &	43.52 &	39.84 &	24.50 &	31.80 &	67.08 & 55.00 &	31.87 &	51.62 &	41.35 \\ 
        $\uparrow$\ours + DPO  & Dynamic ($\delta_{kv}=0.85$) & 49.5\% &   27.99 &	45.29 &	40.46 & 25.84 &	33.40 & 65.51 & 55.20 &	32.73 & 50.28 & 41.86 \\ 
 \midrule
        SmolLM360M-Inst & Base Model & 100\% &  32.76 &	56.10	&52.74&	24.37	&37.00	&70.51	&55.40	&33.59	&54.06 & \textbf{46.28} \\ 
        \hdashline
        $\uparrow$\ours & Random ($r_{kv}=288$)  & 50\% &  22.10	& 36.36	&26.72	&23.22	&25.00	&55.60	&55.20	&23.25	&48.38 & 35.09 \\
        $\uparrow$\ours + DPO & Random ($r_{kv}=288$)  & 50\% &  22.61	&35.61	&27.12	&23.19	&26.60	&56.26	&54.80	&23.64	&49.57 & 35.49 \\
        $\uparrow$\ours & Fixed ($r_{kv}=288$) & 50\% &   33.11	&54.55	&50.77	&23.26	&36.60	&71.06	&55.40	&31.96	&54.14 &	45.65 \\ 
        $\uparrow$\ours + DPO & Fixed ($r_{kv}=288$) & 50\% &  34.39	&55.77	&51.93	&25.68	&38.60	&69.37	&55.20	&33.88	&53.35  &	46.46 \\ 
        $\uparrow$\ours  & Dynamic ($\delta_{kv}=0.88$) & 49.3\% &  32.85	&54.38	&50.96	&23.47	&37.40	&71.11	&55.40	&31.67	&54.62 &	45.76 \\ 
        $\uparrow$\ours + DPO  & Dynamic ($\delta_{kv}=0.88$) & 49.3\% &   34.64	&54.97	&51.75	&25.45	&38.80	&69.59	&55.20	&33.88	&54.06
 &	\textbf{46.48} \\ 
 \midrule
        Llama3.2-1B-Inst & Base Model & 100\% &  37.97 &	63.51 & 60.77 & 46.09 & 35.00 & 74.37 & 60.20 & 38.09 & 59.67 & \textbf{52.85} \\
        \hdashline
        $\uparrow$\ours & Random ($r_{kv}=512$)  & 53.1\% & 35.32	& 60.48&	54.03&	27.77&	35.20&	71.98&	55.80&	33.88&	55.01&	47.72 \\
        $\uparrow$\ours + DPO & Random ($r_{kv}=512$)  & 53.1\% & 38.99 & 62.71	& 56.20 & 28.04 &	36.80 &	73.39 &	56.40 & 36.27 & 56.20 &	49.44 \\
        $\uparrow$\ours & Fixed ($r_{kv}=512$) & 53.1\% & 36.95 &	63.89 &	58.88 &	43.40 &	36.00 &	74.16 &	58.20 &	37.32 &	60.30 &	52.12 \\ 
        $\uparrow$\ours + DPO & Fixed ($r_{kv}=512$) & 53.1\% &  39.93 &	63.89 &	60.73 &	42.39 &	37.80 &	74.92 &	58.80 &	40.77 &	60.54 &	\textbf{53.31} \\ 
        $\uparrow$\ours  & Dynamic ($\delta_{kv}=0.95$) & 54.7\% &   37.12 &	63.64 &	58.87 &	43.26 &	34.40 &	73.72 &	60.00 &	37.51 &	60.22 &	52.08 \\ 
        $\uparrow$\ours + DPO  & Dynamic ($\delta_{kv}=0.95$) & 54.7\% &   41.21 &	64.86 &	60.96 &	42.86 &	37.60 &	74.43 &	58.60 &	39.23 &	58.33 &	53.12 \\
\midrule
        Llama3.2-3B-Inst & Base Model & 100\%   & 45.90 &	67.76 &	70.36 &	60.46 &	36.20 &	75.57 &	69.60 &	40.77 &	67.17 & \textbf{59.31} \\ 
        \hdashline
        $\uparrow$\ours & Random ($r_{kv}=816$)  & 43\%   & 41.64 & 67.34 & 65.11 & 46.97 & 36.40 & 75.24 & 61.6 & 39.04 & 63.69 & 55.23 \\
        $\uparrow$\ours + DPO & Random ($r_{kv}=816$)   & 43\%   & 44.80 & 69.78 & 67.01 & 47.37 & 38.80 & 76.01 & 62.60 & 40.00 & 64.80 & 56.80 \\
        $\uparrow$\ours & Fixed ($r_{kv}=816$) & 43\%   &  43.09 &	67.76 &	69.54 &	56.96 &	37.00 &	75.84 &	66.00 &	41.34 &	67.48 &	58.33 \\ 
        $\uparrow$\ours + DPO & Fixed ($r_{kv}=816$) & 43\%    & 48.21 &	70.45 &	72.24 &	57.42 &	38.40 &	76.55 &	66.80 &	46.22 &	68.59  &	\textbf{60.54} \\ 
        $\uparrow$\ours  & Dynamic ($\delta_{kv}=0.95$) & 43\%   & 42.75 &	66.41 &	69.59 &	57.47 &	36.80 &	75.46 & 67.60 &	42.30 &	68.03 &	58.49 \\ 
        $\uparrow$\ours + DPO  & Dynamic ($\delta_{kv}=0.95$) & 43\%   &  48.46 &	69.99 &	72.26 &	57.73 &	39.40 & 75.79 &	68.40 &	46.32 & 65.90
        &	60.47 \\ 
        \bottomrule

    \end{tabular}
    }
    \caption{{\small{Zero-shot evaluation of self-distilled \ours with different initialization methods (random, SVD with fixed/dynamic rank selection) and base models on the LM Harness Eval benchmark across nine tasks: ARC-Challenge (ARC), ARC-Easy (ARE), HellaSwag (HS), MMLU, OpenBookQA (OBQA), PIQA, PubMedQA (PBMD), RACE (RA), and WinoGrande (WG). ($\uparrow$ denotes upcycling the "Base Model".)}}}
    \label{tab:main_results}
    \vspace{-15pt}
\end{table*}

\vspace{-5pt}
\subsection{Results}

Similar to MambaInLlama~\cite{wang2024mamba}, we adopt the LM Harness Eval benchmark ~\cite{gao2023} (branch big-refactor) to perform zero-shot evaluation on 9 different tasks: ARC-Challenge (ARC)~\cite{clark2018think}, ARC-Easy (ARE)~\cite{clark2018think},HellaSwag (HS)~\cite{zellers2019hellaswag}, MMLU (MM)~\cite{hendrycks2020measuring}, OpenBookQA (OBQA) ~\cite{mihaylov2018can}, PIQA~\cite{bisk2020piqa},  PubMedQA (PBMD) ~\cite{jin2019pubmedqa}, and RACE (RA)~\cite{lai2017race}, WinoGrande (WG) ~\cite{sakaguchi2021winogrande}. 

\vspace{-5pt}

\paragraph{Self-distillation Evaluation} Table \ref{tab:main_results} shows the benchmark performance of our proposed \ours when we use the base models {themselves} as the teacher model, which we refer to as self-distillation. We evaluate three different initialization settings: (i) Fixed rank selection with random initialization, (ii) Fixed rank selection with SVD initialization, and (iii) Dynamic rank selection with SVD initialization. {For all experiments, we set $d_{qk}=d_{r}=32$ and adjust $r_q$, $r_{kv}$, $\delta_q$, and $\delta_{kv}$ accordingly to achieve around $50\%$ KV cache compression while ensuring the total number of parameters after the MLA upcycling remain roughly the same.}  %For the fixed rank selection scenario, we set $r_q=854$, $r_{kv}=512$, and $d_{qk}=d_{r}=32$ such that the total number of parameters after the MLA upcycling remain roughly the same. For the dynamic rank selection case, we apply a threshold of $0.95$ for both $r_q$ and $r_{kv}$ so that the number of parameters aligns with other setups. We investigate two MLA layer upcycling strategies: upcycling $100\%$ of layers to MLA and upcycling $50\%$ of layers to MLA. For the $100\%$ upcycling strategy, we replace all GQA modules in the base model with MLA. In this scenario, the proposed \ours model uses only $53.1\%$ of the KV cache size for fixed rank selection and $54.7\%$ for dynamic rank selection. For the $50\%$ upcycling strategy, we replace GQA modules in layers 1, 3, 5, 7, 8, 10, 12, and 14. This brings us $78.1\%$ KV cache size for the fixed rank selection and $78\%$ for dynamic rank selection. 
{For each scenario, we evaluate training with the full dataset (6.8B tokens for SFT and 0.2B tokens for DPO) and it is observed that DPO significant boosts the distillation performance.} 
%For each strategy, we evaluate training with the full dataset (6.8B tokens) and half dataset (3.4B tokens).  
%It is evident that for fixed rank selection schemes, SVD initialization significantly enhances distillation performance compared to random initialization, yielding an $8\%$ improvement for $100\%$ MLA and $3\%$ improvement for $50\%$ MLA. 
{For fixed rank selection schemes, it is evident that SVD initialization significantly enhances distillation performance compared to random initialization, yielding $22.8\%$ and $30.91\%$ improvements for SmolLM models and $8.1\%$ and $6.5\%$ improvements for Llama 3.2 models. Such observation demonstrates the necessity of applying our SVD initialization method to inherit the knowledge from the pre-trained targe models. For dynamic rank selection schemes, we observe similar performance as the fix rank selection in most experiments, which demonstrates its effectiveness. Note that our \ours+ DPO could always achieve better performance than the pre-trained base models with only $43\%-54.7\%$ KV cache sizes.}%Such advantage is also illustrated by the loss curves shown in Figure \ref{fig:first loss plot}. 
%With dynamic rank selection, the average score is further improved while maintaining a similar KV cache size, achieving the best performance after DPO when training with the full dataset. In general, using more training data boosts the performance. When training with the full dataset, our \ours could greatly outperform the pre-trained base model after DPO. For $100\%$ MLA, we get an improvement of $0.65$ for fix rank selection and an improvement of $0.77$ for dynamic rank selection. For $50\%$ MLA, we could get a even higher improment of $0.89$ for fix rank selection and $0.94$ for dynamic rank selection. 
%Even only trained with the half dataset, \ours does not experience significant performance degradation and could still provide similar or better performance compared to the base model. 
\vspace{-5pt}
\paragraph{Extreme KV Cache Compression with Larger Teacher}
\label{sec:extreme_compression}
Table~\ref{tab:results_compression} presents the impact of KV Cache compression on model accuracy across various benchmarks. {For more details, please refer to Appendix~\ref{app:extreme_kv} and Table~\ref{tab:results_kv}.} {We adopt Llama3.2-1B-Instruct as the base model and progressively reduce the KV cache size of our \ours from $53.1\%$ to \mingyu{$7.81\%$}. With the same base model as our teacher, we observe a consistent accuracy drop across most evaluation tasks, which demonstrates the trade-off between reducing memory consumption and maintaining model accuracy.} 
%As we progressively reduce the KV-cache size—from 53.1\% to 28.1\% and ultimately to 15.6\%—we observe a consistent accuracy drop across most evaluation tasks. 

{However, our results reveal that such performance degradation from extreme KV cache compression can be mitigate if we utilize larger teacher models such as Llama3.2-3B-Instruct and Llama3.1-8B-Instruct. For instance, when reducing the KV cache to $15.6\%$, using Llama3.1-8B-Inst as the teacher recovers $1.56$ of the average score ($52.94$ vs. $51.38$) when trained with half of the dataset (3.6B tokens). With the larger teacher, our \ours achieves even better performance than the pre-trained base model with only $15.6\%$ KV cache size and 3.6B training tokens. As we increase the training tokens to 7B, we could even push the KV cache compression ratio to $9.4\%$ without significant performance degradation compared to the pre-trained base model ($52.47$ vs. $52.85$). These results highlight the effectiveness of leveraging larger teachers and preference tuning to resolve the adverse effects of extreme KV cache compression while maintaining strong accuracy across multiple NLP benchmarks.}

\begin{table*}[t]
     \setlength\extrarowheight{2pt}
    \centering
    \scalebox{0.64}{
    \begin{tabular}{llcccccccccccc}
        \toprule
        Model and Setting & Teacher & Param & Tokens & ARC & ARE & HS & MMLU & OBQA & PIQA & PBMD & RA & WG & Avg. \\
        \midrule
        Llama3.2-1B-Inst & - & 1.24B & - & 37.97 &	63.51 & 60.77 & 46.09 & 35.00 & 74.37 & 60.20 & 38.09 & 59.67 & 52.85 \\
        \midrule
        \multicolumn{14}{c}{100\% MLA Layers ($r_{kv}=512$, $r_q = 864$, $d_{qk}=32$) - KV Size: \textbf{53.1\%}}
        \\
        \midrule
        % $\uparrow$\ours & Llama3.2-1B-Inst & 1.23B & 3.4B & 37.37 &	64.35 &	58.36 &	42.03 &	35.00 &	73.61 &	57.40 &	37.03 &	59.51 & 51.63 \\ 
        
        $\uparrow$\ours + DPO & Llama3.2-1B-Inst & 1.23B & 3.6B &  39.93 &	63.51 &	60.52 &	41.58 &	37.20 &	73.99 &	59.80 &	40.48 &	60.38 &	53.04 \\ 
        
        % $\uparrow$\ours  & Llama3.2-3B-Inst & 1.23B & 3.4B &  37.71	& 65.19 &	58.84 &	43.13 &	36.20 &	73.45 &	58.20 &	37.89 &	59.67 &	 52.25 \\ 
        
        $\uparrow$\ours + DPO  & Llama3.2-3B-Inst & 1.23B &3.6B & 42.75 &	64.81 &	62.04 &	43.88 &	37.40 &	73.72 &	59.20 &	41.44 &	61.48 &	54.08 \\ 
        
        % $\uparrow$\ours & Llama3.2-8B-Inst & 1.23B & 3.4B &  39.51	& 67.38 &	60.41 &	43.18 &	38.40 &	73.94 &	60.40 &	38.28 &	61.72 & 53.69\\ 
        \rowcolor{gray!15}
        $\uparrow$\ours + DPO & Llama3.1-8B-Inst & 1.23B & 3.6B &  44.03 &	68.86 &	63.49 &	43.81 &	37.40 &	73.94 &	61.40 &	41.82 &	61.40 &	\textbf{55.13} \\
        \midrule
        \multicolumn{14}{c}{100\% MLA Layers ($r_{kv}=256$, $r_q = 1184$, $d_{qk}=32$) - KV Size: \textbf{28.1\%}}
        \\
        \midrule
        % $\uparrow$\ours & Llama3.2-1B-Inst & 1.23B & 3.4B & 37.54 &	62.84 &	56.89 &	41.22 &	33.6 &	73.12 &	55.4 &	36.46 &	59.19 & 50.70 \\ 
        
        $\uparrow$\ours + DPO & Llama3.2-1B-Inst & 1.23B & 3.6B &  40.02 &	63.26 &	58.74 &	39.79 &	36.40 &	72.80 &	55.60 &	40.19 &	60.38 &	51.91 \\ 
        
        % $\uparrow$\ours  & Llama3.2-3B-Inst & 1.23B & 3.4B &  36.35 &	63.51 &	57.09 &	41.30 &	35.00 &	73.07 &	56.80 &	36.46 &	60.14 &	 51.08 \\ 
        
        $\uparrow$\ours + DPO  & Llama3.2-3B-Inst & 1.23B &3.6B &  40.70 &	64.35 &	60.10 &	41.77 &	37.20 &	73.83 &	57.80 &	39.23 &	61.17 &	52.91 \\ 
        
        % $\uparrow$\ours & Llama3.2-8B-Inst & 1.23B & 3.4B &  38.14 &	65.45 &	58.70 &	41.15 &	36.20 &	73.67 &	59.00 &	36.17 &	60.62  & 52.12\\ 
        \rowcolor{gray!15}
        $\uparrow$\ours + DPO & Llama3.1-8B-Inst & 1.23B & 3.6B &  41.98 &	66.46 &	61.33 &	41.78 &	37.20 &	74.27 &	59.00 &	40.00 &	60.69 &	\textbf{53.63} \\
        \midrule
        
        \multicolumn{14}{c}{100\% MLA Layers ($r_{kv}=128$, $r_q = 1344$, $d_{qk}=32$) - KV Size: \textbf{15.6\%}}
        \\
        \midrule
        % $\uparrow$\ours & Llama3.2-1B-Inst & 1.23B & 3.4B & 36.52 &	61.41 &	55.37 &	38.02 &	34.60 &	72.52 &	56.00 &	35.60 &	58.56 & 49.84 \\ 
        
        $\uparrow$\ours + DPO & Llama3.2-1B-Inst & 1.23B & 3.6B &  39.16 &	61.83 &	57.27 &	37.85 &	36.20 &	73.45 &	56.40 &	40.19 &	60.06 &	51.38 \\ 
        
        % $\uparrow$\ours  & Llama3.2-3B-Inst & 1.23B & 3.4B &  36.26 &	61.95 &	55.84 &	39.28 &	35.40 &	71.76 &	57.60 &	35.89 &	59.27&	 50.36 \\ 
        
        $\uparrow$\ours + DPO  & Llama3.2-3B-Inst & 1.23B &3.6B &  39.42 &	62.88 &	58.41 &	39.45 &	37.20 &	73.39 &	58.00 &	39.71 &	59.75 &	52.02 \\ 
        
        % $\uparrow$\ours & Llama3.2-8B-Inst & 1.23B & 3.4B &  36.35 &	64.60 &	57.32 &	38.25 &	37.00 &	73.45 &	60.40 &	35.22 &	58.25  & 51.20\\ 
        \rowcolor{gray!15}
        $\uparrow$\ours + DPO & Llama3.1-8B-Inst & 1.23B & 3.6B &  41.30 &	65.61 &	59.64 &	39.47 &	37.60 &	74.27 &	59.20 &	39.52 &	59.83 &	\textbf{52.94} \\

        \hdashline
        % $\uparrow$\ours & Llama3.2-1B-Inst & 1.23B & 6.8B & 37.54 &	62.21 &	56.36 &	39.67 & 35.40 &	73.23 &	55.60 &	35.31 &	58.33 & 50.41 \\ 
        
        $\uparrow$\ours + DPO & Llama3.2-1B-Inst & 1.23B & 7B &  40.10 &	62.88 &	58.17 &	39.70 &	37.80 &	73.50 &	56.60 & 39.33 &	59.67 &	51.97 \\ 
        
        % $\uparrow$\ours  & Llama3.2-3B-Inst & 1.23B & 6.8B &  35.58 &	63.51 &	56.71 &	41.38 &	35.80 &	72.80 &	57.20 &	35.89 &	58.56 &	 50.83 \\ 
        
        $\uparrow$\ours + DPO  & Llama3.2-3B-Inst & 1.23B &7B &  39.33 &	64.86 &	58.92 &	41.86 &	37.40 &	73.83 &	58.80 &	39.71 &	59.59 &	52.70 \\ 
        
        % $\uparrow$\ours & Llama3.2-8B-Inst & 1.23B & 6.8B &  38.65 &	66.88 &	58.46 &	42.01 &	34.80 &	73.67 &	60.00 &	36.46 &	59.12 & 52.23\\ 
        \rowcolor{gray!15}
        $\uparrow$\ours + DPO & Llama3.1-8B-Inst & 1.23B & 7B &  42.49 &	67.13 &	60.58 &	42.51 &	36.60 &	73.99 &	59.40 &	40.38 &	59.43  &	\textbf{53.61} \\
        
        \midrule
        \multicolumn{14}{c}{100\% MLA Layers ($r_{kv}=64$, $r_q = 1424$, $d_{qk}=32$) - KV Size: \textbf{9.4\%}}
        \\
        \midrule
        % $\uparrow$\ours & Llama3.2-1B-Inst & 1.23B & 6.8B & 37.12 &	61.32 &	54.46 &	34.89 &	35.60 &	72.36 &	56.80 &	35.22 &	57.30  & 49.45 \\ 
        
        $\uparrow$\ours + DPO & Llama3.2-1B-Inst & 1.23B & 7B &  39.16 &	62.63 &	56.04 &	34.90 &	36.40 &	72.85 &	56.40 &	37.70 &	58.33  & 50.49 \\ 
        
        % $\uparrow$\ours  & Llama3.2-3B-Inst & 1.23B & 6.8B &  35.07 &	61.95 &	54.95 &	38.61 &	35.20 &	72.09 &	57.40 &	35.98 &	58.25 &	49.94 \\ 
        
        $\uparrow$\ours + DPO  & Llama3.2-3B-Inst & 1.23B &7B &  37.97 &	63.55 &	56.95 &	37.54 &	35.40 &	72.74 &	57.00 &	38.66 &	59.27 &	51.01 \\ 
        
        % $\uparrow$\ours & Llama3.2-8B-Inst & 1.23B & 6.8B &  36.09 &	65.07 &	57.01 &	38.60 &	35.80 &	72.96 &	58.00 &	35.98 &	59.98  & 51.05\\ 
        \rowcolor{gray!15}
        $\uparrow$\ours + DPO & Llama3.1-8B-Inst & 1.23B & 7B &  39.85 &	67.13 &	58.45 &	38.51 &	37.40 &	73.83 &	58.00 &	39.81 &	59.27 &	\textbf{52.47} \\

        \midrule
        \multicolumn{14}{c}{100\% MLA Layers ($r_{kv}=48$, $r_q = 1440$, $d_{qk}=32$) - KV Size: {\textbf{7.81\%}}}
        \\
        \midrule
        % $\uparrow$\ours & Llama3.2-1B-Inst & 1.23B & 6.8B & 36.77 &	60.61 &	53.51 &	32.44 &	33.40 &	72.20 &	56.60 &	34.55 &	58.33  & 48.71 \\ 
        
        $\uparrow$\ours + DPO & Llama3.2-1B-Inst & 1.23B & 7B &  38.48 &	61.66 &	55.32 &	30.62 &	35.20 &	72.36 &	56.60 &	37.99 &	59.43  & 49.74 \\ 
        
        % $\uparrow$\ours  & Llama3.2-3B-Inst & 1.23B & 6.8B & 33.70 &	61.32 &	54.11 &	35.96 &	34.60 &	71.27 &	56.00 &	35.22 &	58.48 &	48.96 \\ 
        
        $\uparrow$\ours + DPO  & Llama3.2-3B-Inst & 1.23B &7B &  36.18 &	62.21 &	55.82 &	36.41 &	35.60 &	72.03 &	57.00 &	38.09 &	60.06  &	50.38 \\ 
        
        % $\uparrow$\ours & Llama3.2-8B-Inst & 1.23B & 6.8B &  36.35 &	64.60 &	55.50 &	36.65 &	34.60 &	72.31 &	57.80 &	35.79 &	58.25  & 50.21\\ 
        \rowcolor{gray!15}
        $\uparrow$\ours + DPO & Llama3.1-8B-Inst & 1.23B & 7B &  37.71 &	65.32 &	57.32 &	36.27 &	36.80 &	72.96 &	58.20 &	38.76 &	58.80 &	\textbf{51.35} \\

        \bottomrule
    \end{tabular}
    }
    \vspace{-5pt}
    \caption{\small{Impact of KV-cache compression and teacher model size on performance. Reducing the KV-cache size lowers accuracy, but larger teacher models help recover performance. DPO further improves alignment and accuracy. ($\uparrow$ denotes upcycling the base model.)}}
    \label{tab:results_compression}
    \vspace{-10pt}
\end{table*}

\vspace{-5pt}

\paragraph{Scalability to Larger Models}
\mehdi{We have extended X-EcoMLA to two \mingyu{8B-parameter} models (Llama3-8B and Llama3.1-8B). The results in Table~\ref{exp:larger} demonstrate that even under aggressive KV cache compression, X-EcoMLA maintains performance very close to the full-scale baseline. Notably, at KV-size \mingyu{10.94\%}, performance remains nearly on par with the full model.}

\begin{table}[htbp]
\centering
 \scalebox{0.7}{
\begin{tabular}{lccccccccccc}
\toprule    
\textbf{Model} & \textbf{KV-size} & \textbf{Avg.} & \textbf{ARC} & \textbf{ARE} & \textbf{HS} & \textbf{MM} & \textbf{OBQA} & \textbf{PIQ} & \textbf{PM} & \textbf{RA} & \textbf{WG} \\
\midrule
Llama3-8B-Inst (Base) & 100\% & 65.78 & 56.66 & 81.61 & 75.81 & 63.82 & 42.60 & 78.62 & 75.00 & 46.03 & 71.90 \\
 \rowcolor{gray!15}
\textbf{ $\uparrow$\ours (Ours; $r_{kv}=256$)} & 15.63\% & 65.16 & 54.69 & 81.02 & 75.69 & 59.20 & 44.40 & 77.91 & 74.80 & 48.04 & 70.72 \\
Llama3.1-8B-Inst (Base) & 100\% & 66.63 & 54.86 & 79.55 & 79.23 & 68.13 & 43.00 & 80.90 & 75.40 & 44.69 & 73.88 \\
 \rowcolor{gray!15}
\textbf{ $\uparrow$\ours (Ours; $r_{kv}=160$)} & 10.94\% & 65.85 & 57.17 & 80.35 & 77.57 & 60.13 & 43.00 & 79.16 & 76.20 & 47.85 & 71.19 \\
\bottomrule
\end{tabular}
}
\caption{\small{Performance of X-EcoMLA with 8B models under aggressive KV cache compression} }
\label{exp:larger}
\end{table}

\begin{figure}[t]
\centering
  \includegraphics[width=0.7\columnwidth]{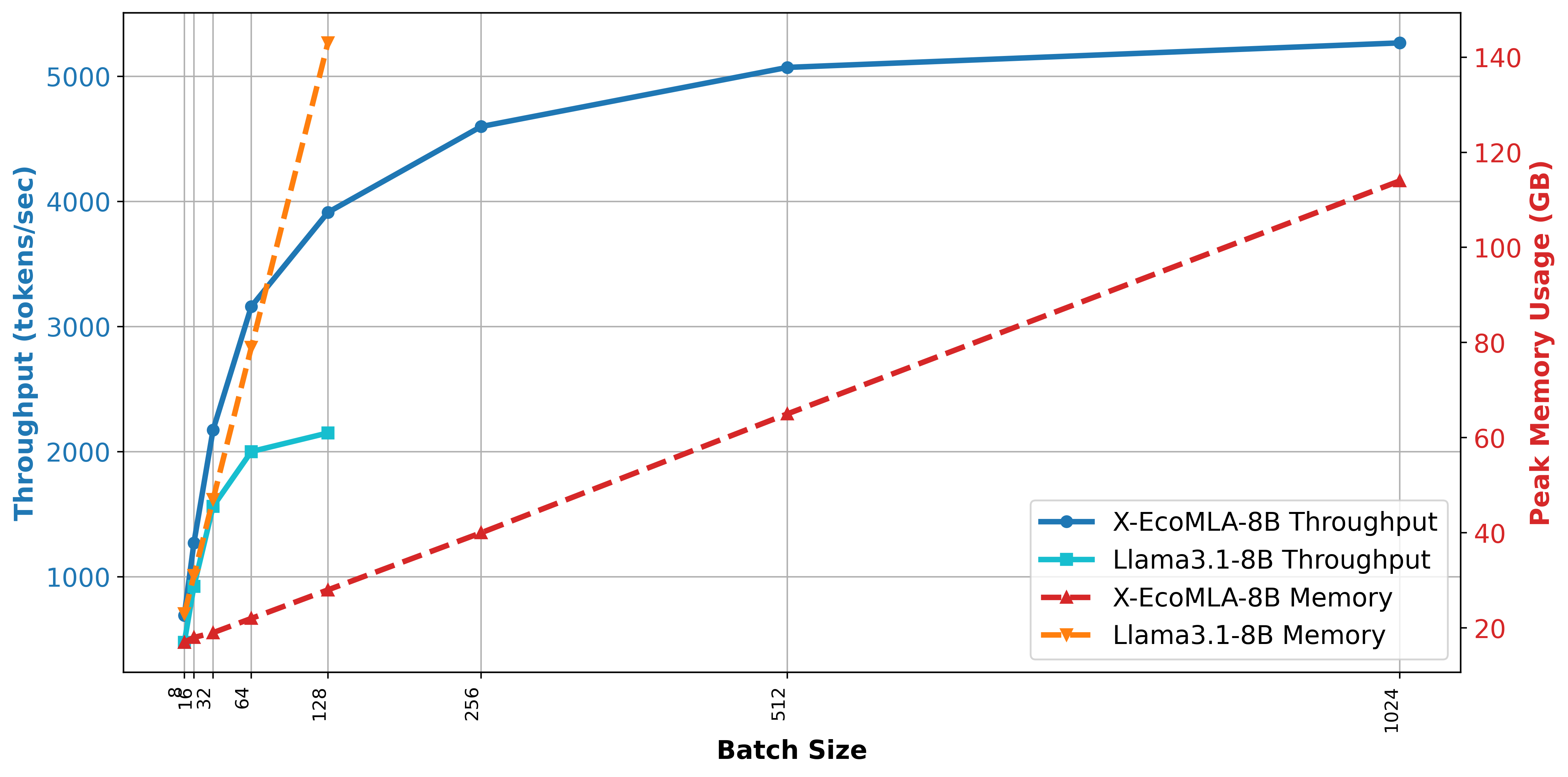}
  \caption{\small{System-level inference performance for Llama3.1-8B and X-EcoMLA-8B on 8$\times$AMD MI300 GPUs. X-EcoMLA enables higher throughput and drastically reduced memory usage across batch sizes. Llama3.1-8B runs out-of-memory (OOM) beyond batch size 128, while X-EcoMLA scales smoothly up to batch size 1024. } }
  \label{fig:inference}
\end{figure}

\paragraph{System-Level Inference Metrics}
\mehdi{We evaluate system-level performance in terms of throughput (sequences/sec) and peak GPU memory usage (GB) for both the baseline model (Llama3.1-8B) and our proposed model, X-EcoMLA-8B ($\mingyu{r_{kv}}=128$, $10.67\times$ KV compression), across a range of batch sizes. All experiments are conducted on identical hardware (single AMD MI300 GPU) under consistent settings.
As shown in Fig.~\ref{fig:inference}, X-EcoMLA-8B achieves approximately $1.7\times$ to $2\times$ higher throughput than the baseline across all batch sizes. %While Llama3.1-8B runs out of memory (OOM) beyond a batch size of 128, X-EcoMLA-8B scales smoothly up to batch size 1024 without encountering OOM. 
In terms of memory efficiency, X-EcoMLA-8B substantially reduces peak memory consumption. For instance, at batch size 128, Llama3.1-8B consumes 143 GB of memory and fails to run larger batches, whereas X-EcoMLA-8B requires only 28 GB—representing a 5$\times$ reduction.
These results demonstrate that X-EcoMLA-8B not only maintains strong model accuracy but also delivers significant system-level improvements—achieving higher throughput and dramatically lower memory usage. This makes it well-suited for latency-sensitive and memory-constrained deployment scenarios.}

\paragraph{Comparison with Existing Post‐Training Low-Rank Methods}
We conducted experiments to compare our solution with two of the most recent SOTA methods — MHA2MLA~\cite{ji2025towards} and PALU~\cite{chang2025palu} — and to clarify how X-EcoMLA improves upon these approaches.

\subparagraph{\textbf{1. Comparison with MHA2MLA Baseline Setup}}

\mehdi{ To more comprehensively evaluate the effectiveness of \ours, we compare it against MHA2MLA~\cite{ji2025towards} under both continual pretraining and supervised fine-tuning (SFT) settings using the SmolLM 1.7B model. As SmolLM is originally built with MHA-based attention, this evaluation also demonstrates that X-EcoMLA can be seamlessly integrated into existing MHA-based architectures.

In the continual pretraining setting, we evaluate X-EcoMLA on the SmolLM 1.7B \textit{base} model using a 12.5\% KV cache size, following the same 6B-token training budget as the released MHA2MLA checkpoint\footnote{\url{https://huggingface.co/fnlp/SmolLM-1B7-MLA-d\_kv\_8}}. We compare four configurations: (1) the full-attention baseline (100\% \mingyu{KV cache}), (2) MHA2MLA continually pretrained, (3) X-EcoMLA pretrained without a teacher, and (4) X-EcoMLA with self-distillation. Table~\ref{res:mha2mla-all} (top) summarizes the results where X-EcoMLA without a teacher achieves an average score of 51.94, slightly outperforming MHA2MLA (51.69) by +0.25. When using self-distillation, the accuracy improves to 52.87, reducing the gap to the full-attention baseline.

In the supervised fine-tuning setting, we use identical SFT data and compare performance under two KV compression ratios: 12.5\% and 50\%. We use SmolLM 1.7B-Instruct as both student and teacher for \ours, and follow the joint-SVD + $S_{higher}$ variant for MHA2MLA. As shown in Table~\ref{res:mha2mla-all} (bottom), X-EcoMLA outperforms MHA2MLA across both compression levels. At 12.5\% KV size, X-EcoMLA achieves 49.34 vs. 48.19 (+1.15), and at 50\%, it reaches 50.15 vs. 49.79 (+0.36).

We attribute this performance gain to a key architectural difference: the RoPE (Rotary Position Embedding) design. MHA2MLA stores separate Key-RoPE vectors per head. Under a fixed budget (e.g., 32 dimensions per token), each head receives only $\frac{32}{\mingyu{n_h}}$ RoPE dimensions. In contrast, \ours adopts the unified RoPE approach used in DeepSeek MLA, where all heads share a single Key-RoPE vector. This allows each head to fully leverage all 32 dimensions, offering $8\times$ more positional encoding capacity in an 8-head setup—crucial for preserving performance under aggressive compression.}

\begin{table}[htbp]
\centering
\scalebox{0.7}{
\begin{tabular}{lccccccccccc}
\toprule
\textbf{Method} & \textbf{KV Size} & \textbf{Avg. Acc.} & \textbf{ARC} & \textbf{ARE} & \textbf{HS} & \textbf{MM} & \textbf{OBQA} & \textbf{PIQA} & \textbf{PM} & \textbf{RA} & \textbf{WG} \\
\midrule

SmolLM 1.7B (Base) & 100\% & 54.67 & 46.42 & 73.48 & 65.74 & 27.73 & 42.00 & 76.06 & 62.60 & 37.03 & 60.93 \\
MHA2MLA \cite{ji2025towards} & 12.5\% & 51.69 & 41.55 & 69.57 & 61.43 & 24.63 & 39.00 & 74.70 & 60.20 & 35.69 & 58.41 \\
 \rowcolor{gray!15}
 $\uparrow$\ours-pretrain ($r_{kv}=480$) & 12.5\% & 51.94 & 40.19 & 69.36 & 62.52 & 23.79 & 40.00 & 75.30 & 61.40 & 35.89 & 59.04 \\
 \rowcolor{gray!15}
 $\uparrow$\ours ($r_{kv}=480$) & 12.5\% & 52.87 & 42.41 & 72.14 & 62.84 & 25.55 & 41.40 & 75.19 & 61.40 & 36.27 & 58.64 \\
 \midrule
SmolLM 1.7B-Ins (Base) & 100\% & 50.49 & 37.71 & 62.96 & 60.81 & 25.86 & 39.80 & 73.50 & 60.80 & 36.17 & 56.83 \\
MHA2MLA (SFT) & 12.5\% & 48.19 & 35.07 & 60.94 & 56.18 & 23.36 & 36.40 & 72.36 & 57.20 & 34.83 & 57.38 \\
\rowcolor{gray!15}
$\uparrow$\ours (SFT, $r_{kv}=480$) & 12.5\% & 49.34 & 37.29 & 62.96 & 59.05 & 23.68 & 38.00 & 72.52 & 59.80 & 34.64 & 56.12 \\
MHA2MLA (SFT) & 50\% & 49.79 & 37.63 & 62.29 & 59.60 & 24.09 & 38.20 & 74.05 & 60.40 & 34.35 & 57.46 \\
\rowcolor{gray!15}
$\uparrow$\ours (SFT, $r_{kv}=2016$) & 50\% & 50.15 & 37.97 & 64.06 & 60.23 & 24.89 & 39.00 & 73.34 & 60.20 & 34.93 & 56.75 \\
\bottomrule
\end{tabular}
}
\caption{\small{Comparison between \ours and MHA2MLA a under both continual pretraining (top) and supervised fine-tuning (bottom) settings on the SmolLM 1.7B model.  } }
\label{res:mha2mla-all}
\end{table}

\subparagraph{\textbf{2. Comparison with PALU}}
\mehdi{PALU~\cite{chang2025palu} represents a recent state-of-the-art approach for low-rank KV-cache compression. It decomposes the linear projection layers into low-rank matrices, caches the compressed intermediate states, and reconstructs the full keys and values on the fly during inference. To evaluate the effectiveness of \ours, we compare X-EcoMLA against PALU using the Llama3-8B-Instruct model.

Following the experimental protocol described in Section 4.2 of the PALU paper, we evaluate zero-shot performance on the used subset of the LM Harness Eval benchmark using PALU's best-performing configurations: G-LRD and J-LRD, both operating at a 50\% KV compression ratio. Reported PALU results are taken directly from their paper.  
The results, summarized in Table~\ref{res:palu}, show that X-EcoMLA achieves an average accuracy of 67.34 at only 15.63\% KV cache size. This performance is nearly on par with the full model (67.87) and clearly outperforms both PALU-J-LRD (66.19) and PALU-G-LRD (64.45), with respective gains of +1.15 and +2.89 points—despite using approximately $3\times$ less KV storage. }

\begin{table}[htbp]
\centering
\scalebox{0.85}{
\begin{tabular}{lcccccccc}
\toprule
\textbf{Model} & \textbf{KV-Size} & \textbf{Avg Acc.} & \textbf{ARC} & \textbf{ARE} & \textbf{HS} & \textbf{OBQA} & \textbf{PIQA} & \textbf{WG} \\
\midrule
Llama3-8B-Inst (Base) & 100\% & 67.87 & 56.66 & 81.61 & 75.81 & 42.60 & 78.62 & 71.90 \\
PALU-J-LRD   & 50\% & 66.19 & 51.96 & 79.63 & 73.20 & 43.40 & 76.50 & 72.45 \\
PALU-G-LRD   & 50\% & 64.45 & 48.99 & 76.30 & 70.36 & 42.60 & 76.06 & 72.38 \\
\rowcolor{gray!15}
 $\uparrow$\ours (Ours) & 15.63\% & 67.34 & 54.69 & 81.02 & 75.69 & 44.40 & 77.53 & 70.72 \\
\bottomrule
\end{tabular}
}
\caption{\small{Comparison between X-EcoMLA and PALU baselines on Llama3-8B. Despite operating at a significantly lower KV size (15.63\%), \ours matches or outperforms 50\%-KV SOTA methods.}}
\label{res:palu}
\end{table}

\paragraph{Expanded Discussion and Experimental Details} 
\mehdi{The Appendix provides additional context and supporting results that complement the main paper. In Section~\ref{related_work}, we present a more detailed discussion of related work on KV cache compression, training-based memory-efficient architectures, and upcycling techniques. Section~\ref{algorithm} outlines the pseudocode for our proposed SVD-based initialization strategy for MLA layers. Hyperparameter for both fixed and dynamic rank selection are reported in Section~\ref{hpo}, helping to guide practical configurations. Section~\ref{supp_res} includes supplementary evaluations: long-context benchmarks on LongBench (Section~\ref{supp_res:long_context}); comparisons with H2O as an alternative KV cache compression method (Section~\ref{supp_res:kv_cach}); hybrid MLA variants that combine attention and MLA layers (Section~\ref{supp_res:hybrid_mla}); and detailed results from our extreme KV cache compression experiments (Section~\ref{app:extreme_kv}). In Section~\ref{sec:ablation}, we present ablation studies, analyzing the impact of distillation loss versus direct supervision (Section~\ref{sec:ce}), the role of LayerNorm (Section~\ref{ablation:ln}), and the trade-off between larger teacher models and longer training (Section~\ref{ablation:data_vs_teacher}). Collectively, these sections provide deeper insights into the design choices and empirical robustness of \ours across diverse settings and compression budgets.}

\section{Conclusion}

In this work, we introduced \ours, a lightweight post-training adaptation approach that enables the upcycling of pre-trained Transformer attention into an efficient MLA or hybrid variant. By leveraging dark knowledge from a well-trained teacher model and employing SVD-based initialization, our method significantly reduces KV cache memory requirements without sacrificing model performance. 
{Our results show that using an 8B teacher model allows us to compress the KV cache size of the Llama3.2-1B-Instruct baseline by 6.4× while preserving 100\% of its average score across multiple tasks on the LM Harness Evaluation benchmark. This is achieved with only 3.6B training tokens and about 70 GPU hours on AMD MI300 GPUs. Alternatively, we can  compress KV cache by 10.6× using ~7B training tokens over approximately 140 GPU hours, while maintaining 99.8\% of the average score.}
% Our experiments demonstrate that \ours can achieve up to 6.4× KV cache compression while maintaining 100\% accuracy on the LM Harness Evaluation benchmark. Moreover, this adaptation is achieved with minimal computational cost, requiring only 3.4B training tokens and 9 hours of training on 8 AMD MI300 GPUs, compared to the extensive compute demands of full MLA pre-training. 
These findings highlight the potential of \ours as a practical and scalable solution for integrating MLA into existing LLMs, paving the way for more memory-efficient and deployable models without the need for costly retraining from scratch.

%



% \section{Ethics Statement}
% Authors can add an optional ethics statement to the paper. 
% For papers that touch on ethical issues, this section will be evaluated as part of the review process. The ethics statement should come at the end of the paper. It does not count toward the page limit, but should not be more than 1 page. 

% \bibliographystyle{colm2024_conference}
% \bibliography{colm2024_conference}

% \bibliographystyle{unsrt}
% \bibliographystyle{natbib}
\bibliographystyle{colm2024_conference}
\bibliography{colm2024_conference}

\begin{thebibliography}{47}
\providecommand{\natexlab}[1]{#1}
\providecommand{\url}[1]{\texttt{#1}}
\expandafter\ifx\csname urlstyle\endcsname\relax
  \providecommand{\doi}[1]{doi: #1}\else
  \providecommand{\doi}{doi: \begingroup \urlstyle{rm}\Url}\fi

\bibitem[Ainslie et~al.(2023)Ainslie, Lee-Thorp, de~Jong, Zemlyanskiy, Lebr{\'o}n, and Sanghai]{ainslie2023gqa}
Joshua Ainslie, James Lee-Thorp, Michiel de~Jong, Yury Zemlyanskiy, Federico Lebr{\'o}n, and Sumit Sanghai.
\newblock Gqa: Training generalized multi-query transformer models from multi-head checkpoints.
\newblock \emph{arXiv preprint arXiv:2305.13245}, 2023.

\bibitem[Arora et~al.(2024)Arora, Eyuboglu, Zhang, Timalsina, Alberti, Zinsley, Zou, Rudra, and R{\'e}]{arora2024simple}
Simran Arora, Sabri Eyuboglu, Michael Zhang, Aman Timalsina, Silas Alberti, Dylan Zinsley, James Zou, Atri Rudra, and Christopher R{\'e}.
\newblock Simple linear attention language models balance the recall-throughput tradeoff.
\newblock \emph{arXiv preprint arXiv:2402.18668}, 2024.

\bibitem[Beltagy et~al.(2020)Beltagy, Peters, and Cohan]{beltagy2020longformer}
Iz~Beltagy, Matthew~E Peters, and Arman Cohan.
\newblock Longformer: The long-document transformer.
\newblock \emph{arXiv preprint arXiv:2004.05150}, 2020.

\bibitem[Bick et~al.(2024)Bick, Li, Xing, Kolter, and Gu]{bick2024transformers}
Aviv Bick, Kevin~Y Li, Eric~P Xing, J~Zico Kolter, and Albert Gu.
\newblock Transformers to ssms: Distilling quadratic knowledge to subquadratic models.
\newblock \emph{arXiv preprint arXiv:2408.10189}, 2024.

\bibitem[Bisk et~al.(2020)Bisk, Zellers, Gao, Choi, et~al.]{bisk2020piqa}
Yonatan Bisk, Rowan Zellers, Jianfeng Gao, Yejin Choi, et~al.
\newblock Piqa: Reasoning about physical commonsense in natural language.
\newblock In \emph{Proceedings of the AAAI conference on artificial intelligence}, volume~34, pp.\  7432--7439, 2020.

\bibitem[Brown et~al.(2020)Brown, Mann, Ryder, Subbiah, Kaplan, Dhariwal, Neelakantan, Shyam, Sastry, Askell, et~al.]{brown2020language}
Tom Brown, Benjamin Mann, Nick Ryder, Melanie Subbiah, Jared~D Kaplan, Prafulla Dhariwal, Arvind Neelakantan, Pranav Shyam, Girish Sastry, Amanda Askell, et~al.
\newblock Language models are few-shot learners.
\newblock \emph{Advances in neural information processing systems}, 33:\penalty0 1877--1901, 2020.

\bibitem[Chang et~al.(2025)Chang, Lin, Lin, Chen, Hu, Wang, Huang, Ceze, Abdelfattah, and Wu]{chang2025palu}
Chi-Chih Chang, Wei-Cheng Lin, Chien-Yu Lin, Chong-Yan Chen, Yu-Fang Hu, Pei-Shuo Wang, Ning-Chi Huang, Luis Ceze, Mohamed~S Abdelfattah, and Kai-Chiang Wu.
\newblock Palu: Kv-cache compression with low-rank projection.
\newblock In \emph{The Thirteenth International Conference on Learning Representations}, 2025.

\bibitem[Chen et~al.(2024)Chen, Qadri, Wen, Jain, Kirchenbauer, Zhou, and Goldstein]{chen2024genqa}
Jiuhai Chen, Rifaa Qadri, Yuxin Wen, Neel Jain, John Kirchenbauer, Tianyi Zhou, and Tom Goldstein.
\newblock Genqa: Generating millions of instructions from a handful of prompts.
\newblock \emph{arXiv preprint arXiv:2406.10323}, 2024.

\bibitem[Chowdhery et~al.(2022)Chowdhery, Narang, Devlin, Bosma, Mishra, Roberts, Barham, Chung, Sutton, Gehrmann, Schuh, Shi, Tsvyashchenko, Maynez, Rao, Barnes, Tay, Shazeer, Prabhakaran, Reif, Du, Hutchinson, Pope, Bradbury, Austin, Isard, Gur-Ari, Yin, Duke, Levskaya, Ghemawat, Dev, Michalewski, Garcia, Misra, Robinson, Fedus, Zhou, Ippolito, Luan, Lim, Zoph, Spiridonov, Sepassi, Dohan, Agrawal, Omernick, Dai, Pillai, Pellat, Lewkowycz, Moreira, Child, Polozov, Lee, Zhou, Wang, Saeta, Diaz, Firat, Catasta, Wei, Meier-Hellstern, Eck, Dean, Petrov, and Fiedel]{chowdhery2022palm}
Aakanksha Chowdhery, Sharan Narang, Jacob Devlin, Maarten Bosma, Gaurav Mishra, Adam Roberts, Paul Barham, Hyung~Won Chung, Charles Sutton, Sebastian Gehrmann, Parker Schuh, Kensen Shi, Sasha Tsvyashchenko, Joshua Maynez, Abhishek Rao, Parker Barnes, Yi~Tay, Noam Shazeer, Vinodkumar Prabhakaran, Emily Reif, Nan Du, Ben Hutchinson, Reiner Pope, James Bradbury, Jacob Austin, Michael Isard, Guy Gur-Ari, Pengcheng Yin, Toju Duke, Anselm Levskaya, Sanjay Ghemawat, Sunipa Dev, Henryk Michalewski, Xavier Garcia, Vedant Misra, Kevin Robinson, Liam Fedus, Denny Zhou, Daphne Ippolito, David Luan, Hyeontaek Lim, Barret Zoph, Alexander Spiridonov, Ryan Sepassi, David Dohan, Shivani Agrawal, Mark Omernick, Andrew~M. Dai, Thanumalayan~Sankaranarayana Pillai, Marie Pellat, Aitor Lewkowycz, Erica Moreira, Rewon Child, Oleksandr Polozov, Katherine Lee, Zongwei Zhou, Xuezhi Wang, Brennan Saeta, Mark Diaz, Orhan Firat, Michele Catasta, Jason Wei, Kathy Meier-Hellstern, Douglas Eck, Jeff Dean, Slav Petrov, and Noah Fiedel.
\newblock Palm: Scaling language modeling with pathways, 2022.

\bibitem[Clark et~al.(2018)Clark, Cowhey, Etzioni, Khot, Sabharwal, Schoenick, and Tafjord]{clark2018think}
Peter Clark, Isaac Cowhey, Oren Etzioni, Tushar Khot, Ashish Sabharwal, Carissa Schoenick, and Oyvind Tafjord.
\newblock Think you have solved question answering? try arc, the ai2 reasoning challenge.
\newblock \emph{arXiv preprint arXiv:1803.05457}, 2018.

\bibitem[Cui et~al.(2023)Cui, Yuan, Ding, Yao, Zhu, Ni, Xie, Liu, and Sun]{cui2023ultrafeedback}
Ganqu Cui, Lifan Yuan, Ning Ding, Guanming Yao, Wei Zhu, Yuan Ni, Guotong Xie, Zhiyuan Liu, and Maosong Sun.
\newblock Ultrafeedback: Boosting language models with high-quality feedback, 2023.

\bibitem[Dao \& Gu(2024)Dao and Gu]{dao2024transformers}
Tri Dao and Albert Gu.
\newblock Transformers are ssms: Generalized models and efficient algorithms through structured state space duality.
\newblock \emph{arXiv preprint arXiv:2405.21060}, 2024.

\bibitem[Dong et~al.(2024)Dong, Fu, Diao, Byeon, Chen, Mahabaleshwarkar, Liu, Keirsbilck, Chen, Suhara, Lin, Kautz, and Molchanov]{dong2024hymbahybridheadarchitecturesmall}
Xin Dong, Yonggan Fu, Shizhe Diao, Wonmin Byeon, Zijia Chen, Ameya~Sunil Mahabaleshwarkar, Shih-Yang Liu, Matthijs~Van Keirsbilck, Min-Hung Chen, Yoshi Suhara, Yingyan Lin, Jan Kautz, and Pavlo Molchanov.
\newblock Hymba: A hybrid-head architecture for small language models, 2024.
\newblock URL \url{https://arxiv.org/abs/2411.13676}.

\bibitem[Gao et~al.(2023)Gao, Tow, Biderman, Black, DiPofi, Foster, Golding, Hsu, McDonell, Muennighoff, Phang, Reynolds, Tang, Thite, Wang, Wang, and Zou]{gao2023}
Leo Gao, Jonathan Tow, Stella Biderman, Sid Black, Anthony DiPofi, Charles Foster, Laurence Golding, Jeffrey Hsu, Kyle McDonell, Niklas Muennighoff, Jason Phang, Laria Reynolds, Eric Tang, Anish Thite, Ben Wang, Kevin Wang, and Andy Zou.
\newblock A framework for few-shot language model evaluation.
\newblock 2023.

\bibitem[Grattafiori et~al.(2024)Grattafiori, Dubey, Jauhri, Pandey, Kadian, Al-Dahle, Letman, Mathur, Schelten, Vaughan, et~al.]{grattafiori2024Llama_31}
Aaron Grattafiori, Abhimanyu Dubey, Abhinav Jauhri, Abhinav Pandey, Abhishek Kadian, Ahmad Al-Dahle, Aiesha Letman, Akhil Mathur, Alan Schelten, Alex Vaughan, et~al.
\newblock The llama 3 herd of models.
\newblock \emph{arXiv preprint arXiv:2407.21783}, 2024.

\bibitem[Gu \& Dao()Gu and Dao]{gu_mamba_nodate}
Albert Gu and Tri Dao.
\newblock Mamba: {Linear}-{Time} {Sequence} {Modeling} with {Selective} {State} {Spaces}.

\bibitem[Guo et~al.(2025)Guo, Yang, Zhang, Song, Zhang, Xu, Zhu, Ma, Wang, Bi, et~al.]{guo2025deepseek}
Daya Guo, Dejian Yang, Haowei Zhang, Junxiao Song, Ruoyu Zhang, Runxin Xu, Qihao Zhu, Shirong Ma, Peiyi Wang, Xiao Bi, et~al.
\newblock Deepseek-r1: Incentivizing reasoning capability in llms via reinforcement learning.
\newblock \emph{arXiv preprint arXiv:2501.12948}, 2025.

\bibitem[He et~al.(2024)He, Khattar, Prenger, Korthikanti, Yan, Liu, Fan, Aithal, Shoeybi, and Catanzaro]{he2024upcycling}
Ethan He, Abhinav Khattar, Ryan Prenger, Vijay Korthikanti, Zijie Yan, Tong Liu, Shiqing Fan, Ashwath Aithal, Mohammad Shoeybi, and Bryan Catanzaro.
\newblock Upcycling large language models into mixture of experts.
\newblock \emph{arXiv preprint arXiv:2410.07524}, 2024.

\bibitem[Hendrycks et~al.(2020)Hendrycks, Burns, Basart, Zou, Mazeika, Song, and Steinhardt]{hendrycks2020measuring}
Dan Hendrycks, Collin Burns, Steven Basart, Andy Zou, Mantas Mazeika, Dawn Song, and Jacob Steinhardt.
\newblock Measuring massive multitask language understanding.
\newblock \emph{arXiv preprint arXiv:2009.03300}, 2020.

\bibitem[Ji et~al.(2025)Ji, Guo, Wu, Guo, Shen, Chen, Qiu, Zhang, and Gui]{ji2025towards}
Tao Ji, Bin Guo, Yuanbin Wu, Qipeng Guo, Lixing Shen, Zhan Chen, Xipeng Qiu, Qi~Zhang, and Tao Gui.
\newblock Towards economical inference: Enabling deepseek's multi-head latent attention in any transformer-based llms.
\newblock \emph{arXiv preprint arXiv:2502.14837}, 2025.

\bibitem[Jin et~al.(2019)Jin, Dhingra, Liu, Cohen, and Lu]{jin2019pubmedqa}
Qiao Jin, Bhuwan Dhingra, Zhengping Liu, William~W Cohen, and Xinghua Lu.
\newblock Pubmedqa: A dataset for biomedical research question answering.
\newblock \emph{arXiv preprint arXiv:1909.06146}, 2019.

\bibitem[Kang et~al.(2024)Kang, Zhang, Kundu, Jeong, Liu, Krishna, and Zhao]{kang2024gear}
Hao Kang, Qingru Zhang, Souvik Kundu, Geonhwa Jeong, Zaoxing Liu, Tushar Krishna, and Tuo Zhao.
\newblock Gear: An efficient kv cache compression recipefor near-lossless generative inference of llm.
\newblock \emph{arXiv e-prints}, pp.\  arXiv--2403, 2024.

\bibitem[Komatsuzaki et~al.(2022)Komatsuzaki, Puigcerver, Lee-Thorp, Ruiz, Mustafa, Ainslie, Tay, Dehghani, and Houlsby]{komatsuzaki2022sparse}
Aran Komatsuzaki, Joan Puigcerver, James Lee-Thorp, Carlos~Riquelme Ruiz, Basil Mustafa, Joshua Ainslie, Yi~Tay, Mostafa Dehghani, and Neil Houlsby.
\newblock Sparse upcycling: Training mixture-of-experts from dense checkpoints.
\newblock \emph{arXiv preprint arXiv:2212.05055}, 2022.

\bibitem[Lai et~al.(2017)Lai, Xie, Liu, Yang, and Hovy]{lai2017race}
Guokun Lai, Qizhe Xie, Hanxiao Liu, Yiming Yang, and Eduard Hovy.
\newblock Race: Large-scale reading comprehension dataset from examinations.
\newblock \emph{arXiv preprint arXiv:1704.04683}, 2017.

\bibitem[Lian et~al.(2023)Lian, Goodson, Pentland, Cook, Vong, and "Teknium"]{OpenOrca}
Wing Lian, Bleys Goodson, Eugene Pentland, Austin Cook, Chanvichet Vong, and "Teknium".
\newblock Openorca: An open dataset of gpt augmented flan reasoning traces.
\newblock \url{https://https://huggingface.co/Open-Orca/OpenOrca}, 2023.

\bibitem[Lieber et~al.(2024)Lieber, Lenz, Bata, Cohen, Osin, Dalmedigos, Safahi, Meirom, Belinkov, Shalev-Shwartz, Abend, Alon, Asida, Bergman, Glozman, Gokhman, Manevich, Ratner, Rozen, Shwartz, Zusman, and Shoham]{lieber2024jamba}
Opher Lieber, Barak Lenz, Hofit Bata, Gal Cohen, Jhonathan Osin, Itay Dalmedigos, Erez Safahi, Shaked Meirom, Yonatan Belinkov, Shai Shalev-Shwartz, Omri Abend, Raz Alon, Tomer Asida, Amir Bergman, Roman Glozman, Michael Gokhman, Avashalom Manevich, Nir Ratner, Noam Rozen, Erez Shwartz, Mor Zusman, and Yoav Shoham.
\newblock Jamba: A hybrid transformer-mamba language model, 2024.

\bibitem[Liu et~al.(2024{\natexlab{a}})Liu, Feng, Wang, Wang, Liu, Zhao, Dengr, Ruan, Dai, Guo, et~al.]{liu2024deepseek}
Aixin Liu, Bei Feng, Bin Wang, Bingxuan Wang, Bo~Liu, Chenggang Zhao, Chengqi Dengr, Chong Ruan, Damai Dai, Daya Guo, et~al.
\newblock Deepseek-v2: A strong, economical, and efficient mixture-of-experts language model.
\newblock \emph{arXiv preprint arXiv:2405.04434}, 2024{\natexlab{a}}.

\bibitem[Liu et~al.(2024{\natexlab{b}})Liu, Feng, Xue, Wang, Wu, Lu, Zhao, Deng, Zhang, Ruan, et~al.]{liu2024deepseek-v3}
Aixin Liu, Bei Feng, Bing Xue, Bingxuan Wang, Bochao Wu, Chengda Lu, Chenggang Zhao, Chengqi Deng, Chenyu Zhang, Chong Ruan, et~al.
\newblock Deepseek-v3 technical report.
\newblock \emph{arXiv preprint arXiv:2412.19437}, 2024{\natexlab{b}}.

\bibitem[Mihaylov et~al.(2018)Mihaylov, Clark, Khot, and Sabharwal]{mihaylov2018can}
Todor Mihaylov, Peter Clark, Tushar Khot, and Ashish Sabharwal.
\newblock Can a suit of armor conduct electricity? a new dataset for open book question answering.
\newblock \emph{arXiv preprint arXiv:1809.02789}, 2018.

\bibitem[of~Artificial Intelligence~(BAAI)(2024)]{infinity_struct}
Beijing~Academy of~Artificial Intelligence~(BAAI).
\newblock Infinity instruct.
\newblock \url{https://huggingface.co/datasets/BAAI/Infinity-Instruct}, 2024.

\bibitem[OpenAI(2023)]{openai2023gpt4}
OpenAI.
\newblock Gpt-4 technical report, 2023.

\bibitem[Poli et~al.(2024)Poli, Thomas, Nguyen, Ponnusamy, Deiseroth, Kersting, Suzuki, Hie, Ermon, R{\'e}, et~al.]{poli2024mechanistic}
Michael Poli, Armin~W Thomas, Eric Nguyen, Pragaash Ponnusamy, Bj{\"o}rn Deiseroth, Kristian Kersting, Taiji Suzuki, Brian Hie, Stefano Ermon, Christopher R{\'e}, et~al.
\newblock Mechanistic design and scaling of hybrid architectures.
\newblock \emph{arXiv preprint arXiv:2403.17844}, 2024.

\bibitem[Qin et~al.(2024)Qin, Li, Sun, Sun, Shen, Han, Wei, Lv, Luo, Qiao, and Zhong]{qin2024transnormerllmfasterbetterlarge}
Zhen Qin, Dong Li, Weigao Sun, Weixuan Sun, Xuyang Shen, Xiaodong Han, Yunshen Wei, Baohong Lv, Xiao Luo, Yu~Qiao, and Yiran Zhong.
\newblock Transnormerllm: A faster and better large language model with improved transnormer, 2024.
\newblock URL \url{https://arxiv.org/abs/2307.14995}.

\bibitem[Sakaguchi et~al.(2021)Sakaguchi, Bras, Bhagavatula, and Choi]{sakaguchi2021winogrande}
Keisuke Sakaguchi, Ronan~Le Bras, Chandra Bhagavatula, and Yejin Choi.
\newblock Winogrande: An adversarial winograd schema challenge at scale.
\newblock \emph{Communications of the ACM}, 64\penalty0 (9):\penalty0 99--106, 2021.

\bibitem[Shazeer(2019)]{shazeer2019fast}
Noam Shazeer.
\newblock Fast transformer decoding: One write-head is all you need.
\newblock \emph{arXiv preprint arXiv:1911.02150}, 2019.

\bibitem[Shi et~al.(2024)Shi, Zhang, Yao, Li, and Zhao]{shi2024keep}
Luohe Shi, Hongyi Zhang, Yao Yao, Zuchao Li, and Hai Zhao.
\newblock Keep the cost down: A review on methods to optimize llm's kv-cache consumption.
\newblock \emph{arXiv preprint arXiv:2407.18003}, 2024.

\bibitem[Sun et~al.(2024)Sun, Dong, Zhu, Huang, Wang, Ma, Zhang, Wang, and Wei]{sun2024you}
Yutao Sun, Li~Dong, Yi~Zhu, Shaohan Huang, Wenhui Wang, Shuming Ma, Quanlu Zhang, Jianyong Wang, and Furu Wei.
\newblock You only cache once: Decoder-decoder architectures for language models.
\newblock \emph{Advances in Neural Information Processing Systems}, 37:\penalty0 7339--7361, 2024.

\bibitem[Teknium(2023)]{OpenHermes2_5}
Teknium.
\newblock Openhermes 2.5: An open dataset of synthetic data for generalist llm assistants, 2023.
\newblock URL \url{https://huggingface.co/datasets/teknium/OpenHermes-2.5}.

\bibitem[Wang(2024)]{ultrafeedback_armorm}
Junxiong Wang.
\newblock Llama3 ultrafeedback-armorm dataset.
\newblock \url{https://huggingface.co/datasets/JunxiongWang/llama3-ultrafeedback-armorm}, 2024.

\bibitem[Wang et~al.(2024{\natexlab{a}})Wang, Paliotta, May, Rush, and Dao]{wang2024mamba}
Junxiong Wang, Daniele Paliotta, Avner May, Alexander Rush, and Tri Dao.
\newblock The mamba in the llama: Distilling and accelerating hybrid models.
\newblock \emph{Advances in Neural Information Processing Systems}, 37:\penalty0 62432--62457, 2024{\natexlab{a}}.

\bibitem[Wang et~al.(2024{\natexlab{b}})Wang, Jin, Yu, and Zhang]{wang2024model}
Zheng Wang, Boxiao Jin, Zhongzhi Yu, and Minjia Zhang.
\newblock Model tells you where to merge: Adaptive kv cache merging for llms on long-context tasks.
\newblock \emph{arXiv preprint arXiv:2407.08454}, 2024{\natexlab{b}}.

\bibitem[Xiao et~al.(2023)Xiao, Tian, Chen, Han, and Lewis]{xiao2023efficient}
Guangxuan Xiao, Yuandong Tian, Beidi Chen, Song Han, and Mike Lewis.
\newblock Efficient streaming language models with attention sinks.
\newblock \emph{arXiv preprint arXiv:2309.17453}, 2023.

\bibitem[Yang et~al.(2024)Yang, Wang, Shen, Panda, and Kim]{yang2024gated}
Songlin Yang, Bailin Wang, Yikang Shen, Rameswar Panda, and Yoon Kim.
\newblock Gated linear attention transformers with hardware-efficient training, 2024.

\bibitem[Zellers et~al.(2019)Zellers, Holtzman, Bisk, Farhadi, and Choi]{zellers2019hellaswag}
Rowan Zellers, Ari Holtzman, Yonatan Bisk, Ali Farhadi, and Yejin Choi.
\newblock Hellaswag: Can a machine really finish your sentence?
\newblock \emph{arXiv preprint arXiv:1905.07830}, 2019.

\bibitem[Zhang et~al.(2024{\natexlab{a}})Zhang, Ji, Chen, Fu, Miao, Nie, Chen, and Cui]{zhang2024pqcache}
Hailin Zhang, Xiaodong Ji, Yilin Chen, Fangcheng Fu, Xupeng Miao, Xiaonan Nie, Weipeng Chen, and Bin Cui.
\newblock Pqcache: Product quantization-based kvcache for long context llm inference.
\newblock \emph{arXiv preprint arXiv:2407.12820}, 2024{\natexlab{a}}.

\bibitem[Zhang et~al.(2024{\natexlab{b}})Zhang, Bhatia, Kumbong, and Re]{zhang2024the}
Michael Zhang, Kush Bhatia, Hermann Kumbong, and Christopher Re.
\newblock The hedgehog \& the porcupine: Expressive linear attentions with softmax mimicry.
\newblock In \emph{The Twelfth International Conference on Learning Representations}, 2024{\natexlab{b}}.
\newblock URL \url{https://openreview.net/forum?id=4g02l2N2Nx}.

\bibitem[Zhang et~al.(2023)Zhang, Sheng, Zhou, Chen, Zheng, Cai, Song, Tian, R{\'e}, Barrett, et~al.]{zhang2023h2o}
Zhenyu Zhang, Ying Sheng, Tianyi Zhou, Tianlong Chen, Lianmin Zheng, Ruisi Cai, Zhao Song, Yuandong Tian, Christopher R{\'e}, Clark Barrett, et~al.
\newblock H2o: Heavy-hitter oracle for efficient generative inference of large language models.
\newblock \emph{Advances in Neural Information Processing Systems}, 36:\penalty0 34661--34710, 2023.

\end{thebibliography}

\appendix
page
\section{Appendix}

\subsection{Related Work}
\label{related_work}

\paragraph{KV Cache Management in Transformers}
% Transformers store key-value (KV) vectors for each token and each layer during auto-regressive generation, leading to high memory requirements during inference, especially for long sequences. 
{Several approaches have been proposed to reduce or compress the KV cache size of Transformers}, which can be broadly categorized into training-based and post-training solutions~\cite{shi2024keep}. Training-based methods involve modifying the model architecture and pre-training it, typically yielding better performance, whereas post-training methods are easier to apply and do not require retraining.

A variety of post-training KV cache management solutions have been explored in the literature. One common strategy is KV cache eviction, such as Heavy Hitter ($H_2O$)~\cite{zhang2023h2o}, which defines an eviction policy based on the observation that only a few tokens contribute to the highest attention scores. This method retains the most recent and most significant tokens while discarding the others. Another approach is sliding window attention~\cite{arora2024simple, beltagy2020longformer}, which restricts attention to a fixed number of recent tokens (or predefined patterns) to maintain a bounded KV cache size. Attention Sink~\cite{xiao2023efficient} builds on this by retaining initial tokens in the KV cache to improve performance. Quantization-based KV cache compression~\cite{kang2024gear, zhang2024pqcache} reduces memory usage by storing KVs in a lower-precision format, while KV cache merging~\cite{wang2024model} minimizes information loss by merging KV entries instead of discarding them.
Although post-training solutions are computationally efficient, they often lead to performance degradation due to information loss. In contrast, training-based methods offer a better balance between memory efficiency and model accuracy. This paper focuses on training-based solutions, which we review in the following.
 
\paragraph{Training-based KV Cache Management}
Training-based solutions modify the attention mechanism or replace it with alternative architectures in Transformer models to reduce KV cache memory requirements. For instance, multi-query attention (MQA)\cite{shazeer2019fast} shares keys and values across all attention heads, reducing the KV cache size by a factor of $n$ compared to a multi-head attention (MHA) model with $n$ KV heads. However, sharing a single KV across n query heads can be too restrictive. To address this, grouped-query attention (GQA)\cite{ainslie2023gqa} divides query heads into groups, allowing each group to share a single set of key and value heads, making a balance between memory efficiency and performance.
Another notable approach is YOCO~\cite{sun2024you}, a decode-decoder model that consists of a self-decoder and a cross-decoder module. Instead of storing KV vectors for each layer and token, the self-decoder module provides a shared global KV cache to the cross-decoder layers, significantly reducing memory overhead.
Multi-head latent attention (MLA), introduced in DeepSeek-V2~\cite{liu2024deepseek}, is another KV-cache efficient variation of MHA. MLA reduces KV cache size by projecting input hidden states into a compressed latent space through low-rank projection, leading to a substantial reduction in memory usage. DeepSeek-V2 demonstrated that MLA can outperform standard MHA while maintaining efficiency.

Motivated by MLA’s strong performance and efficiency, we focus on adapting MLA for already pre-trained models. However, training-based solutions typically require full pre-training from scratch or extensive continual training.
% , making them computationally expensive and resource-intensive.
This raises a fundamental question: Can we upcycle pre-trained models to their MLA counterparts without costly retraining? In the following section, we review existing solutions for model upcycling that can be leveraged for this purpose.
 
\paragraph{Upcycling Attention }
In \cite{komatsuzaki2022sparse}, model upcycling is defined as "upgrading an existing model with a relatively small additional computational budget." This term has primarily been used to describe the conversion of dense models into mixture-of-experts (MoE) models in an efficient manner~\cite{komatsuzaki2022sparse, he2024upcycling}. In this paper, we focus on the concept of attention upcycling, which involves adapting pre-trained attention blocks in a Transformer into more efficient forms, such as MLA, without requiring full re-training from scratch.
There are several examples of attention upcycling in the literature. For instance, in GQA, \citet{ainslie2023gqa} propose replacing MHA blocks with GQA and performing light continual pre-training for adaptation. Similarly, Hedgehog\cite{zhang2024the} introduces an upcycling method that converts pre-trained attention into linear attention using knowledge distillation.

A notable line of work focuses on leveraging the duality between Transformer self-attention and alternative architectures. MambaInLlama~\cite{wang2024mamba} demonstrates this by replacing some attention layers in pre-trained models with Mamba layers, initializing them from their corresponding attention layers, and then fine-tuning using end-to-end knowledge distillation. 
% This process transfers quadratic knowledge from Transformers to hybrid sub-quadratic attention-Mamba models.
Similarly, MOHAWK~\cite{bick2024transformers} follows a knowledge distillation-based approach for training hybrid attention-Mamba models. However, MOHAWK differs from MambaInLlama in some aspects: 
(a) It does not initialize the student sub-quadratic model from the Transformer attention layers;
(b) It incorporates intermediate layer distillation in addition to end-to-end distillation.

\subsection{Algorithm}
\label{algorithm}
{Our simple method for initializing the MLA weights using SVD approach applied to the pre-trained attention weights is summarized in the pseudocode in Algorithm~\ref{alg:pseudocode}.}

\begin{algorithm}[t]
\begin{lstlisting}
# MHA weights: W_Q, W_K, W_V
# MLA weights: W_DQ, W_UQ, W_QR, W_DKV, W_UK, W_KR, W_UV

# Initialization of W_DQ, W_UQ, and W_QR
U_q, sigma_q, V_q = svd(W_Q)
W_DQ = U_q
W_UQR_bar = (sigma_q @ V_q).view(r_q, n_h, d_h)
W_UQ = W_UQR_bar[:, :, :d_qk].view(r_q, n_h*d_qk)
W_QR = W_UQR_bar[:, :, -d_r:].view(r_q, n_h*d_r)

# Initialization of W_DKV, W_UK, W_KR, W_UV
U_kv, sigma_kv, V_kv = svd(torch.cat((W_K, W_V), -1))
W_DKV = U_kv
W_K_avg = W_K.view(d, n_h, d_h).mean(1)
W_KR = W_K_avg[:, -d_r:]

W_UKV = sigma_kv @ V_kv
W_UK_bar = W_UKV[:, :d_h*n_h].view(r_kv, n_h, d_h)
W_UK = W_UK_bar[:,:,:d_qk].view(r_kv, n_h*d_qk)
W_UV = W_UKV[:, d_h*n_h:]
\end{lstlisting}
\caption{Python-like pseudocode of the proposed SVD initialization for MLA.}
\label{alg:pseudocode}
\end{algorithm}

\subsection{Hyper-parameter Selection}
\label{hpo}
In Table~\ref{tab:HPO} and ~\ref{tab:HPO_dynamic}, we present the model performance with different hyperparameters for fixed rank selection and dynamic rank selection, respectively. In Table~\ref{tab:HPO}, we evaluate three KV ranks ($r_{kv}=512, 256, 128$) and two head dimensions ($d_{qk}=32, 64$). We adjust $r_q$ accordingly to make sure all configurations have approximately the same number of parameters. The results indicate a significant performance loss as the KV rank $r_{kv}$ decreases. With the same KV rank, $d_{qk}=64$ generally provides better performance. However, such advantage is more obvious with $r_{kv} = 128, 256$ where $r_q$ is relatively large. When $r_{kv}=512$, both head dimensions provides similar performance. In Table~\ref{tab:HPO_dynamic}, we explore two thresholds (90\% and 95\%) for $r_q$ and $r_{kv}$ and two head dimensions ($d_{qk}=32, 64$) for dynamic rank selection. When training with a small portion of the dataset (1.6B), we notice that the performance is mainly influenced by the KV rank $r_{kv}$. Although setting $d_{qk}=64$ leads to more parameters, it does not necessarily translate to performance improvement, even when trained with the full dataset.

\begin{table}[h]
    \centering
    \begin{tabular}{l c c c c c c c}
        \toprule
        \textbf{Configuration} & Param & $r_q$ & $r_{kv}$ & $d_{qk}$ & KV Size & Tokens & \textbf{Avg Score} \\
        \midrule
        \rowcolor{gray!15}
        \multicolumn{8}{c}{Base Model: \textbf{Llama3.2-1B-Inst}}\\
        $\uparrow$\ours +DPO  & 1.23B & 864  & 512 & 32 & 53.1\% & 3.6B & 53.04\\
        $\uparrow$\ours +DPO & 1.23B & 480  & 512 & 64 &  56.3\% & 3.6B & 53.14\\
        $\uparrow$\ours +DPO & 1.23B & 1184  & 256 & 32 &  28.1\% & 3.6B & 51.91\\
        $\uparrow$\ours +DPO & 1.23B & 736  & 256 & 64 &  31.3\% & 3.6B & 52.38\\
        $\uparrow$\ours +DPO & 1.23B & 1344  & 128 & 32 &  15.6\% & 3.6B &51.38 \\
        $\uparrow$\ours +DPO & 1.23B & 864  & 128 & 64 &  18.8\% & 3.6B & 51.60\\       
        \bottomrule
    \end{tabular}
    \caption{\small{Hyperparameter selection for the internal dimensions of the \ours block under a fixed setting with 100\% MLA layers, without LayerNorm, and using an identical teacher model as the base.} }
    \label{tab:HPO}
\end{table}

\begin{table}[h]
    \centering
    \begin{tabular}{l c c c c c c c}
        \toprule
        \textbf{Configuration} & Param & $r_q$ & $r_{kv}$ & $d_{qk}$ & KV Size & Tokens & \textbf{Avg Score} \\
        \midrule
        \rowcolor{gray!15}
        \multicolumn{8}{c}{Base Model: \textbf{Llama3.2-1B-Inst}}\\
$\uparrow$\ours +DPO & 1.22B & 90\% & 90\% & 32 & 42.7\% & 1.6B & 51.26 \\
$\uparrow$\ours +DPO & 1.25B & 90\% & 90\% & 64 & 45.9\% & 1.6B & 51.31 \\
$\uparrow$\ours +DPO & 1.23B & 95\% & 90\% & 32 & 42.7\% & 1.6B & 51.36 \\
$\uparrow$\ours +DPO & 1.27B & 95\% & 90\% & 64 & 45.9\% & 1.6B & 51.21 \\
$\uparrow$\ours +DPO & 1.23B & 90\% & 95\% & 32 & 54.7\% & 1.6B & 52.18 \\
$\uparrow$\ours +DPO & 1.26B & 90\% & 95\% & 64 & 57.9\% & 1.6B & 51.51 \\
$\uparrow$\ours +DPO & 1.24B & 95\% & 95\% & 32 & 54.7\% & 1.6B & 52.40 \\
$\uparrow$\ours +DPO & 1.28B & 95\% & 95\% & 64 & 57.9\% & 1.6B & 52.16 \\
$\uparrow$\ours +DPO & 1.23B & 90\% & 95\% & 32 & 54.7\% & 7.0B & 53.22 \\
$\uparrow$\ours +DPO & 1.26B & 90\% & 95\% & 64 & 57.9\% & 7.0B & 53.23 \\
\bottomrule
    \end{tabular}
    \caption{\small{Hyperparameter selection for the internal dimensions of the \ours block under a dynamic setting with 100\% MLA layers, without LayerNorm, and using an identical teacher model (Llama-3.2-1B) as the base.} }
    \label{tab:HPO_dynamic}
\end{table}

\subsection{Supplementary Results}
\label{supp_res}
\subsubsection{Long Context Evaluations}
\label{supp_res:long_context}
\mehdi{We evaluated our MLA-optimized models on the LongBench benchmark, which covers a range of long-context understanding tasks such as LCC, Qasper, QMSum, Multi-News, and SamSum. Table~\ref{res:long_bench} reports the results under various KV-cache size for both Llama3.2-1B and Llama3.2-3B models. 
Notably, X-EcoMLA 3B models achieves a score of 60.03 on LCC, significantly outperforming the full-sized Llama3.2-3B baseline (52.11), despite using only 43\% of the KV cache. Across other tasks such as Qasper, Multi-News, and SamSum, our compressed models match or even slightly exceed the performance of their full-cache counterparts.
These results indicate that our method scales well to long-sequence scenarios and is particularly effective in memory-constrained environments.}

\begin{table}[htbp]
\centering
 \scalebox{0.7}{
\begin{tabular}{lcccccccc}
\toprule
\textbf{Model} & \textbf{KV-Size} & \textbf{Avg. Acc.} & \textbf{lcc} & \textbf{repobench-p} & \textbf{qasper} & \textbf{qmsum} & \textbf{multi\_news} & \textbf{samsum} \\
\midrule
Llama3.2-1B-Inst (Base) & 100.00\% & 30.805 & 35.47 & 40.12 & 22.92 & 21.65 & 25.68 & 38.99 \\
 \rowcolor{gray!15}
$\uparrow$\ours (ours) & 53.13\% & 30.77 & 38.73 & 40.36 & 21.13 & 20.50 & 25.76 & 38.11 \\
 \rowcolor{gray!15}
 $\uparrow$\ours (ours) & 28.13\% & 30.66 & 38.74 & 40.54 & 21.21 & 20.61 & 25.62 & 37.26 \\
\midrule
Llama3.2-3B-Inst (Base) & 100.00\% & 40.01 & 52.11 & 54.16 & 40.42 & 23.63 & 26.51 & 43.21 \\
 \rowcolor{gray!15}
 $\uparrow$\ours  (ours) & 42.91\% & 39.29 & 60.03 & 56.24 & 29.94 & 21.08 & 27.54 & 40.93 \\
 \rowcolor{gray!15}
 $\uparrow$\ours  (ours) & 25.00\% & 39.11 & 59.59 & 53.94 & 31.75 & 20.93 & 27.19 & 41.26 \\
\bottomrule
\end{tabular}
}
\caption{\small{Long-context evaluation on the LongBench benchmark across varying KV cache sizes. All the {X-EcoMLA} models are trained with Llama3.1-8B-Inst  as the teacher model}}
\label{res:long_bench}
\end{table}

\subsubsection{Comparison with other KV Cache Compression Techniques}
\label{supp_res:kv_cach}
\mehdi{We compare X-EcoMLA with the widely used H2O method~\cite{zhang2023h2o}, using the same base model (Llama3.2-1B-Instruct) and identical KV cache sizes. The evaluation is conducted on the \textbf{lm-eval-hardness} benchmark to assess performance under increasingly aggressive memory constraints.
As shown in Table~\ref{tab:xecomla-vs-h2o}, X-EcoMLA consistently outperforms H2O across all compression levels—both in terms of average accuracy and on most individual tasks. Notably, at a KV size of 9.4\%, X-EcoMLA achieves an average accuracy of \textbf{50.49\%}, compared to \textbf{45.05\%} for H2O, with particularly large gains on ARC, ARE, and PIQA. Even at 6.25\% KV size, X-EcoMLA maintains a strong lead, indicating its robustness under extreme compression.
These results demonstrate that X-EcoMLA achieves significantly better accuracy under the same memory budget, making it a strong candidate for memory-efficient inference. 
}

\begin{table}[htbp]
\centering
\scalebox{0.78}{
\begin{tabular}{lccccccccccc}
\toprule
\textbf{Model} & \textbf{KV-size} & \textbf{Avg. Acc.} & \textbf{ARC} & \textbf{ARE} & \textbf{HS} & \textbf{MM} & \textbf{OBQA} & \textbf{PIQ} & \textbf{PM} & \textbf{RA} & \textbf{WG} \\
\midrule
H2O & 15.6\% & 50.30 & 37.71 & 57.41 & 59.91 & 40.83 & 31.60 & 71.11 & 60.40 & 37.99 & 55.80 \\
\rowcolor{gray!15}
$\uparrow$\ours (ours) & 15.6\% & 51.97 & 40.10 & 62.88 & 58.17 & 39.70 & 37.80 & 73.50 & 56.60 & 39.33 & 59.67 \\
\midrule
H2O & 9.4\% & 45.05 & 30.03 & 43.01 & 57.79 & 33.25 & 29.60 & 64.96 & 58.80 & 36.08 & 51.93 \\
\rowcolor{gray!15}
$\uparrow$\ours (ours) & 9.4\% & 50.49 & 39.16 & 62.63 & 56.04 & 34.90 & 36.40 & 72.85 & 56.40 & 37.70 & 58.33 \\
\midrule
H2O & 6.25\% & 41.30 & 26.54 & 34.68 & 52.75 & 26.95 & 28.60 & 59.03 & 58.60 & 34.26 & 50.28 \\
\rowcolor{gray!15}
$\uparrow$\ours (ours) & 6.25\% & 49.74 & 38.48 & 61.66 & 55.32 & 30.62 & 35.20 & 72.36 & 56.60 & 37.99 & 59.43 \\
\bottomrule
\end{tabular}
}
\caption{\small{Comparison of X-EcoMLA and H2O~\cite{zhang2023h2o} across various KV cache sizes. X-EcoMLA consistently outperforms H2O, especially under aggressive compression.}}
\label{tab:xecomla-vs-h2o}
\end{table}

\subsubsection{Hybrid MLA Models}
\label{supp_res:hybrid_mla}
In this section, we include some supplementary results. Table \ref{tab:results} shows the benchmark performance of our \ours method on Llama3.2-1B-Inst model when we use the same model as teacher.  We evaluate three different initialization settings: (i) Fixed rank selection with random initialization, (ii) Fixed rank selection with SVD initialization, and (iii) Dynamic rank selection with SVD initialization. For the fixed rank selection scenario, we set $r_q=854$, $r_{kv}=512$, and $d_{qk}=d_{r}=32$ such that the total number of parameters after the MLA upcycling remain roughly the same. For the dynamic rank selection case, we apply a threshold of $0.95$ for both $r_q$ and $r_{kv}$ so that the number of parameters aligns with other setups. We investigate two MLA layer upcycling strategies: upcycling $100\%$ of layers to MLA and upcycling $50\%$ of layers to MLA. For the $100\%$ upcycling strategy, we replace all GQA modules in the base model with MLA. In this scenario, the proposed \ours model uses only $53.1\%$ of the KV cache size for fixed rank selection and $54.7\%$ for dynamic rank selection. For the $50\%$ upcycling strategy, we replace GQA modules in layers 1, 3, 5, 7, 8, 10, 12, and 14. This brings us $78.1\%$ KV cache size for the fixed rank selection and $78\%$ for dynamic rank selection. 

For each strategy, we evaluate training with the full dataset (6.8B tokens) and half dataset (3.4B tokens).  
It is evident that for fixed rank selection schemes, SVD initialization significantly enhances distillation performance compared to random initialization, yielding an $8\%$ improvement for $100\%$ MLA and $3\%$ improvement for $50\%$ MLA. 

\begin{table*}[h]
     \setlength\extrarowheight{2pt}
    \centering
    \scalebox{0.64}{
    \begin{tabular}{llcccccccccccc}
        \toprule
        Model and Setting & Init. Method & KV-Size & Tokens & ARC & ARE & HS & MMLU & OBQA & PIQA & PBMD & RA & WG & Avg. \\
        \midrule
        Llama3.2-1B-Inst & Base & 100\% & - & 37.97 &	63.30 & 60.65 & 46.05 & 34.80 & 74.32 & 60.00 & 38.18 & 59.67 & \textbf{52.77} \\
        \midrule
        \multicolumn{14}{c}{100\% MLA Layers- Teacher: Identical to the Base Model}
        \\
        \midrule
        $\uparrow$\ours & Random (512)  & 53.1\% & 6.8B & 35.32	& 60.48&	54.03&	27.77&	35.20&	71.98&	55.80&	33.88&	55.01&	47.72 \\
        $\uparrow$\ours + DPO & Random (512)  & 53.1\% & 7.0B & 38.99 & 62.71	& 56.20 & 28.04 &	36.8 &	73.39 &	56.40 & 36.27 & 56.20 &	49.44 \\
        $\uparrow$\ours & Fixed (512) & 53.1\% & 6.8B & 36.95 &	63.89 &	58.88 &	\textbf{43.40} &	36.00 &	74.16 &	58.20 &	37.32 &	60.30 &	52.12 \\ 
        $\uparrow$\ours + DPO & Fixed (512) & 53.1\% & 7.0B &  40.19 &	63.93 &	60.67 &	42.31 &	\textbf{37.60} &	\textbf{75.03} &	59.20 &	\textbf{40.86} &	\textbf{61.01} &	53.42 \\ 
        $\uparrow$\ours  & Dynamic (95\%) & 54.7\% & 6.8B &   37.12 &	63.64 &	58.87 &	43.26 &	34.40 &	73.72 &	60.00 &	37.51 &	60.22 &	52.08 \\ 
        $\uparrow$\ours + DPO  & Dynamic (95\%) & 54.7\% & 7.0B & \textbf{40.36} &	\textbf{64.31} &	\textbf{60.88} &	42.54 &	36.80 &	74.16 &	\textbf{61.40} &	40.77 &	60.69 &	\textbf{53.54} \\ 
        \midrule
        %\rowcolor{gray!15}
        $\uparrow$\ours & Fixed (512) & 53.1\% & 3.4B &  37.37 &	\textbf{64.35} &	58.36 &	42.03 &	35.00 &	73.61 &	57.40 &	37.03 &	59.51 & 51.63\\ 
        %\rowcolor{gray!15}
        $\uparrow$\ours + DPO & Fixed (512) & 53.1\% & 3.6B &  39.93 &	63.51 &	60.52 &	41.58 &	\textbf{37.20} &	73.99 &	\textbf{59.80} &	\textbf{40.48} &	\textbf{60.38} &	\textbf{53.04} \\
        %\rowcolor{gray!15}
        $\uparrow$\ours  & Dynamic (95\%) & 54.7\% & 3.4B &  37.12 &	63.64 &	58.44 &	\textbf{42.14} &	34.40 &	73.61 &	57.00 &	37.22 &	59.98 &	51.50  \\ 
        %\rowcolor{gray!15}
        $\uparrow$\ours + DPO  & Dynamic (95\%) & 54.7\% & 3.6B & \textbf{40.27} &	62.71 &	\textbf{60.55} &	41.21 &	36.40 &	\textbf{74.16} &	\textbf{59.80} &	39.90 &	60.14 &	52.79 \\
        \midrule
        \multicolumn{14}{c}{50\% MLA Layers, Teacher: Identical to the Base Model}
        \\
        \midrule
        $\uparrow$\ours & Random (512)  & 78.1\% & 6.8B &  36.86 &	62.79 &	57.23 &	38.19 &	36.00 &	73.78 &	56.20 &	36.75 &	58.72 &	50.72 \\
        $\uparrow$\ours + DPO & Random (512)  & 78.1\% & 7.0B&  38.99 &	63.64 &	59.00 &	37.46 &	\textbf{37.60} &	74.59 &	57.00 &	39.14 &	60.46 &	51.99 \\
        $\uparrow$\ours & Fixed (512) & 78.1\% & 6.8B &  37.97 &	63.01 &	59.71 &	\textbf{44.37} &	35.80 &	\textbf{74.86} &	\textbf{60.60} &	38.37 &	59.98 &	52.74 \\ 
        $\uparrow$\ours + DPO & Fixed (512) & 78.1\% & 7.0B & 40.87 &	63.93 &	\textbf{61.95} &	43.39 &	37.20 &	74.48 &	59.80 &	\textbf{40.48} &	60.85 &	53.66  \\        
        $\uparrow$\ours & Dynamic (95\%) & 78\% & 6.8B &  38.48 &	63.85 &	59.78 &	44.27 &	35.60 &	74.54 &	60.40 &	38.18 &	60.85 &	52.88 \\
        $\uparrow$\ours + DPO & Dynamic (95\%) & 78\% &  7.0B & \textbf{41.64}  &	\textbf{64.44} &	61.78 &	43.58 &	36.40 &	74.21 &	60.00 &	\textbf{40.48} &	\textbf{60.93} &	\textbf{53.71} \\
        \midrule
        %\rowcolor{gray!15}
        $\uparrow$\ours & Fixed (512) & 78.1\% & 3.4B &   37.12 &	63.55 &	59.36 &	\textbf{44.16} &	35.20 &	73.94 &	57.60 &	37.70 &	60.77 &	52.16\\ 
        %\rowcolor{gray!15}
        $\uparrow$\ours + DPO & Fixed (512) & 78.1\% & 3.6B & \textbf{40.61} &	\textbf{64.73} &	\textbf{62.06} &	43.51 &	\textbf{37.40} &	73.78 &	59.40 &	40.29 &	\textbf{61.25} &	\textbf{53.67}  \\ 
        %\rowcolor{gray!15}
        $\uparrow$\ours & Dynamic (95\%) & 78\% & 3.4B &   38.05 &	63.09 &	59.37 &	43.60 &	35.00 &	\textbf{74.27} &	59.80 &	36.94 &	60.93 &	52.33 \\
        %\rowcolor{gray!15}
        $\uparrow$\ours + DPO & Dynamic (95\%) & 78\% &  3.6B & 39.93  &	63.64 &	61.76 &	43.33 &	36.40 &	73.83 &	\textbf{61.40} &	\textbf{40.57} &	60.30 &	53.46 \\
        \bottomrule
    \end{tabular}
    }
    \caption{\small{Zero-shot evaluation of MLA variants with different initialization methods (random, SVD with fixed rank selection, and SVD with dynamic rank selection) on the LM Harness Eval benchmark across nine tasks: ARC-Challenge (ARC), ARC-Easy (ARE), HellaSwag (HS), MMLU, OpenBookQA (OBQA), PIQA, PubMedQA (PBMD), RACE (RA), and WinoGrande (WG). ($\uparrow$ denotes upcycling the base model.)}}
    \label{tab:results}
\end{table*}

\subsubsection{More Details on Extreme KV Cache Compression Experiments}
\label{app:extreme_kv}
{In this section, we include more details for Table~\ref{tab:results_compression} results. For each row, we also show the results of SFT training. }

\begin{table*}[h]
     \setlength\extrarowheight{2pt}
    \centering
    \scalebox{0.64}{
    \begin{tabular}{llcccccccccccc}
        \toprule
        Model and Setting & Teacher & Param & Tokens & ARC & ARE & HS & MMLU & OBQA & PIQA & PBMD & RA & WG & Avg. \\
        \midrule
        Llama3.2-1B-Inst & - & 1.24B & - & 37.97 &	63.30 & 60.65 & 46.05 & 34.80 & 74.32 & 60.00 & 38.18 & 59.67 & 52.77 \\
        \midrule
        \multicolumn{14}{c}{100\% MLA Layers ($r_{kv}=512$, $r_q = 864$, $d_{qk}=32$) - KV Size: \textbf{53.1\%}}
        \\
        \midrule
        $\uparrow$\ours & Llama3.2-1B-Inst & 1.23B & 3.4B & 37.37 &	64.35 &	58.36 &	42.03 &	35.00 &	73.61 &	57.40 &	37.03 &	59.51 & 51.63 \\ 
        
        $\uparrow$\ours + DPO & Llama3.2-1B-Inst & 1.23B & 3.6B &  39.93 &	63.51 &	60.52 &	41.58 &	37.20 &	73.99 &	59.80 &	40.48 &	60.38 &	53.04 \\ 
        
        $\uparrow$\ours  & Llama3.2-3B-Inst & 1.23B & 3.4B &  37.71	& 65.19 &	58.84 &	43.13 &	36.20 &	73.45 &	58.20 &	37.89 &	59.67 &	 52.25 \\ 
        
        $\uparrow$\ours + DPO  & Llama3.2-3B-Inst & 1.23B &3.6B & 42.75 &	64.81 &	62.04 &	43.88 &	37.40 &	73.72 &	59.20 &	41.44 &	61.48 &	54.08 \\ 
        
        $\uparrow$\ours & Llama3.2-8B-Inst & 1.23B & 3.4B &  39.51	& 67.38 &	60.41 &	43.18 &	38.40 &	73.94 &	60.40 &	38.28 &	61.72 & 53.69\\ 
        \rowcolor{gray!15}
        $\uparrow$\ours + DPO & Llama3.2-8B-Inst & 1.23B & 3.6B &  44.03 &	68.86 &	63.49 &	43.81 &	37.40 &	73.94 &	61.40 &	41.82 &	61.40 &	\textbf{55.13} \\
        \midrule
        \multicolumn{14}{c}{100\% MLA Layers ($r_{kv}=256$, $r_q = 1184$, $d_{qk}=32$) - KV Size: \textbf{28.1\%}}
        \\
        \midrule
        $\uparrow$\ours & Llama3.2-1B-Inst & 1.23B & 3.4B & 37.54 &	62.84 &	56.89 &	41.22 &	33.6 &	73.12 &	55.4 &	36.46 &	59.19 & 50.70 \\ 
        
        $\uparrow$\ours + DPO & Llama3.2-1B-Inst & 1.23B & 3.6B &  40.02 &	63.26 &	58.74 &	39.79 &	36.40 &	72.80 &	55.60 &	40.19 &	60.38 &	51.91 \\ 
        
        $\uparrow$\ours  & Llama3.2-3B-Inst & 1.23B & 3.4B &  36.35 &	63.51 &	57.09 &	41.30 &	35.00 &	73.07 &	56.80 &	36.46 &	60.14 &	 51.08 \\ 
        
        $\uparrow$\ours + DPO  & Llama3.2-3B-Inst & 1.23B &3.6B &  40.70 &	64.35 &	60.10 &	41.77 &	37.20 &	73.83 &	57.80 &	39.23 &	61.17 &	52.91 \\ 
        
        $\uparrow$\ours & Llama3.2-8B-Inst & 1.23B & 3.4B &  38.14 &	65.45 &	58.70 &	41.15 &	36.20 &	73.67 &	59.00 &	36.17 &	60.62  & 52.12\\ 
        \rowcolor{gray!15}
        $\uparrow$\ours + DPO & Llama3.2-8B-Inst & 1.23B & 3.6B &  41.98 &	66.46 &	61.33 &	41.78 &	37.20 &	74.27 &	59.00 &	40.00 &	60.69 &	\textbf{53.63} \\
        \midrule
        
        \multicolumn{14}{c}{100\% MLA Layers ($r_{kv}=128$, $r_q = 1344$, $d_{qk}=32$) - KV Size: \textbf{15.6\%}}
        \\
        \midrule
        $\uparrow$\ours & Llama3.2-1B-Inst & 1.23B & 3.4B & 36.52 &	61.41 &	55.37 &	38.02 &	34.60 &	72.52 &	56.00 &	35.60 &	58.56 & 49.84 \\ 
        
        $\uparrow$\ours + DPO & Llama3.2-1B-Inst & 1.23B & 3.6B &  39.16 &	61.83 &	57.27 &	37.85 &	36.20 &	73.45 &	56.40 &	40.19 &	60.06 &	51.38 \\ 
        
        $\uparrow$\ours  & Llama3.2-3B-Inst & 1.23B & 3.4B &  36.26 &	61.95 &	55.84 &	39.28 &	35.40 &	71.76 &	57.60 &	35.89 &	59.27&	 50.36 \\ 
        
        $\uparrow$\ours + DPO  & Llama3.2-3B-Inst & 1.23B &3.6B &  39.42 &	62.88 &	58.41 &	39.45 &	37.20 &	73.39 &	58.00 &	39.71 &	59.75 &	52.02 \\ 
        
        $\uparrow$\ours & Llama3.2-8B-Inst & 1.23B & 3.4B &  36.35 &	64.60 &	57.32 &	38.25 &	37.00 &	73.45 &	60.40 &	35.22 &	58.25  & 51.20\\ 
        \rowcolor{gray!15}
        $\uparrow$\ours + DPO & Llama3.2-8B-Inst & 1.23B & 3.6B &  41.30 &	65.61 &	59.64 &	39.47 &	37.60 &	74.27 &	59.20 &	39.52 &	59.83 &	\textbf{52.94} \\

        \hdashline
        $\uparrow$\ours & Llama3.2-1B-Inst & 1.23B & 6.8B & 37.54 &	62.21 &	56.36 &	39.67 & 35.40 &	73.23 &	55.60 &	35.31 &	58.33 & 50.41 \\ 
        
        $\uparrow$\ours + DPO & Llama3.2-1B-Inst & 1.23B & 7B &  40.10 &	62.88 &	58.17 &	39.70 &	37.80 &	73.50 &	56.60 & 39.33 &	59.67 &	51.97 \\ 
        
        $\uparrow$\ours  & Llama3.2-3B-Inst & 1.23B & 6.8B &  35.58 &	63.51 &	56.71 &	41.38 &	35.80 &	72.80 &	57.20 &	35.89 &	58.56 &	 50.83 \\ 
        
        $\uparrow$\ours + DPO  & Llama3.2-3B-Inst & 1.23B &7B &  39.33 &	64.86 &	58.92 &	41.86 &	37.40 &	73.83 &	58.80 &	39.71 &	59.59 &	52.70 \\ 
        
        $\uparrow$\ours & Llama3.2-8B-Inst & 1.23B & 6.8B &  38.65 &	66.88 &	58.46 &	42.01 &	34.80 &	73.67 &	60.00 &	36.46 &	59.12 & 52.23\\ 
        \rowcolor{gray!15}
        $\uparrow$\ours + DPO & Llama3.2-8B-Inst & 1.23B & 7B &  42.49 &	67.13 &	60.58 &	42.51 &	36.60 &	73.99 &	59.40 &	40.38 &	59.43  &	\textbf{53.61} \\
        
        \midrule
        \multicolumn{14}{c}{100\% MLA Layers ($r_{kv}=64$, $r_q = 1424$, $d_{qk}=32$) - KV Size: \textbf{9.4\%}}
        \\
        \midrule
        $\uparrow$\ours & Llama3.2-1B-Inst & 1.23B & 6.8B & 37.12 &	61.32 &	54.46 &	34.89 &	35.60 &	72.36 &	56.80 &	35.22 &	57.30  & 49.45 \\ 
        
        $\uparrow$\ours + DPO & Llama3.2-1B-Inst & 1.23B & 7B &  39.16 &	62.63 &	56.04 &	34.90 &	36.40 &	72.85 &	56.40 &	37.70 &	58.33  & 50.49 \\ 
        
        $\uparrow$\ours  & Llama3.2-3B-Inst & 1.23B & 6.8B &  35.07 &	61.95 &	54.95 &	38.61 &	35.20 &	72.09 &	57.40 &	35.98 &	58.25 &	49.94 \\ 
        
        $\uparrow$\ours + DPO  & Llama3.2-3B-Inst & 1.23B &7B &  37.97 &	63.55 &	56.95 &	37.54 &	35.40 &	72.74 &	57.00 &	38.66 &	59.27 &	51.01 \\ 
        
        $\uparrow$\ours & Llama3.2-8B-Inst & 1.23B & 6.8B &  36.09 &	65.07 &	57.01 &	38.60 &	35.80 &	72.96 &	58.00 &	35.98 &	59.98  & 51.05\\ 
        \rowcolor{gray!15}
        $\uparrow$\ours + DPO & Llama3.2-8B-Inst & 1.23B & 7B &  40.02 &	67.17 &	58.40 &	38.53 &	37.80 &	73.83 &	58.00 &	39.43 &	60.93 &	\textbf{52.68} \\

        \midrule
        \multicolumn{14}{c}{100\% MLA Layers ($r_{kv}=48$, $r_q = 1440$, $d_{qk}=32$) - KV Size: \textbf{6.25\%}}
        \\
        \midrule
        $\uparrow$\ours & Llama3.2-1B-Inst & 1.23B & 6.8B & 36.77 &	60.61 &	53.51 &	32.44 &	33.40 &	72.20 &	56.60 &	34.55 &	58.33  & 48.71 \\ 
        
        $\uparrow$\ours + DPO & Llama3.2-1B-Inst & 1.23B & 7B &  38.48 &	61.66 &	55.32 &	30.62 &	35.20 &	72.36 &	56.60 &	37.99 &	59.43  & 49.74 \\ 
        
        $\uparrow$\ours  & Llama3.2-3B-Inst & 1.23B & 6.8B & 33.70 &	61.32 &	54.11 &	35.96 &	34.60 &	71.27 &	56.00 &	35.22 &	58.48 &	48.96 \\ 
        
        $\uparrow$\ours + DPO  & Llama3.2-3B-Inst & 1.23B &7B &  36.18 &	62.21 &	55.82 &	36.41 &	35.60 &	72.03 &	57.00 &	38.09 &	60.06  &	50.38 \\ 
        
        $\uparrow$\ours & Llama3.2-8B-Inst & 1.23B & 6.8B &  36.35 &	64.60 &	55.50 &	36.65 &	34.60 &	72.31 &	57.80 &	35.79 &	58.25  & 50.21\\ 
        \rowcolor{gray!15}
        $\uparrow$\ours + DPO & Llama3.2-8B-Inst & 1.23B & 7B &  37.71 &	65.32 &	57.32 &	36.27 &	36.80 &	72.96 &	58.20 &	38.76 &	58.80 &	\textbf{51.35} \\

        \bottomrule
    \end{tabular}
    }
    \caption{\small{Impact of KV-cache compression and teacher model size on performance. Reducing the KV-cache size lowers accuracy, but larger teacher models help recover performance. DPO further improves alignment and accuracy. ($\uparrow$ denotes upcycling the base model.)}}
    \label{tab:results_kv}
\end{table*}

\subsection{Ablation Studies}
\label{sec:ablation}

\subsubsection{{Distillation vs. Cross-Entropy}}
\label{sec:ce}
In Table~\ref{tab:loss}, we examine the trade-off between learning from the teacher knowledge (via KL divergence) and direct supervision from the dataset (via cross-entropy loss w.r.t. ground truth) during the SFT distillation stage. {We adopt Llama3.2-1B-Instruct for both our base model and student model, and we use CE and KL to denote the weights for the cross-entropy loss and KL divergence loss.} The results show that relying solely on direct supervision (CE = 1, KL = 0) significantly degrades model performance (48.54 vs. 52.77), underscoring the importance of leveraging teacher knowledge for effective learning.

In contrast, incorporating teacher knowledge—either exclusively or in combination with direct supervision—yields the best results, indicating the importance of teacher-guided learning in maintaining accuracy. Given these insights, we primarily adopt teacher-based learning in our configurations to minimize hyperparameter tuning efforts unless stated otherwise.

\begin{table}[h]
    \centering
    \begin{tabular}{l c c c}
        \toprule
        \textbf{Configuration} & \textbf{CE} & \textbf{KL} & \textbf{Avg Score} \\
        \midrule
        Llama3.2-1B-Inst & - & - & 52.77 \\
        $\uparrow$\ours  & 0 & 1 & 50.84 \\
        $\uparrow$\ours & 1 & 0.01 & 50.93 \\
        $\uparrow$\ours & 1 & 0.05 & 50.71 \\
        $\uparrow$\ours & 1 & 0.1 & 50.98 \\
        $\uparrow$\ours & 1 & 0 & 48.54 \\
        \bottomrule
    \end{tabular}
    \caption{\small{Comparison of different CE and KL loss weightings in the SFT knowledge distillation phase. The experiment utilizes the same teacher as the base model and applies dynamic SVD compression with {$\delta_q=\delta_{kv}=0.95$}, trained on 20\% of the dataset.}}
    \label{tab:loss}
\end{table}

\subsubsection{Impact of LayerNorm}
\label{ablation:ln}
The original MLA module incorporates additional LayerNorm layers between the down- and up-projection {operations}. However, we {observe} that it is beneficial to omit those LayerNorm layers in our proposed \ours {, as evidenced in Table \ref{tab:LN} and the loss curves in Figure \ref{fig:first loss plot} in the Appendix. By removing the intermediate LayerNorm layers, our proposed \ours demonstrates an improved loss convergence. Besides, across various setups (fixed, dynamic) and different training dataset sizes (3.4B, 6.8B) in Table \ref{tab:LN}, the removal of LayerNorm layers consistently leads to performance gains.} 
%This benefit is evident in the loss curves presented in Figure \ref{fig:first loss plot}. By removing the intermediate LayerNorm layers, our proposed scheme demonstrates an improved loss convergence. The advantages of excluding LayerNorm layers are further illustrated in Table \ref{tab:LN}. Across various setups (fixed, dynamic) and different training dataset sizes (3.4B, 6.8B), the removal of LayerNorm layers consistently leads to performance gains. 

\begin{figure}[tbh]
\centering
  \includegraphics[width=0.6\columnwidth]{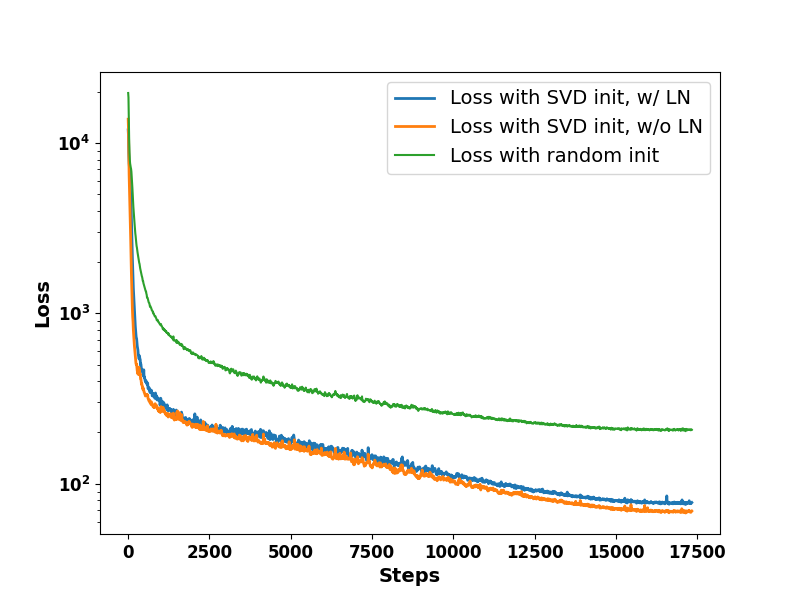}
  \caption{\small{Loss curve comparison between random initialization and SVD initialization w/ and w/o LayerNorm layers. All schemes are trained with fixed KV rank selection and 100\% MLA layers upcycling.} }
  \label{fig:first loss plot}
\end{figure}

\begin{table}[h]
    \centering
    \begin{tabular}{l c c c c c}
        \toprule
        \textbf{Configuration} & Init. Method & LayerNorm & KV Size & Tokens & \textbf{Avg Score} \\
        \midrule
        \rowcolor{gray!15}
        \multicolumn{6}{c}{Base Model: \textbf{Llama3.2-1B-Inst}}\\
        $\uparrow$\ours  & Dynamic ($\delta_{kv}=0.95$) & \cmark  &  54.7\% & 6.8B & 51.68 \\
        $\uparrow$\ours  & Dynamic ($\delta_{kv}=0.95$) & \xmark  &  54.7\% & 6.8B & \textbf{52.08} \\
        $\uparrow$\ours  & Fixed ($r_{kv} = 512$) & \cmark  & 53.1\% & 6.8B & 51.68 \\
        $\uparrow$\ours  & Fixed ($r_{kv} = 512$) & \xmark  & 53.1\% & 6.8B & \textbf{52.12} \\
        $\uparrow$\ours  & Dynamic ($\delta_{kv}=0.95$) & \cmark  &  54.7\% & 3.4B & 51.18 \\
        $\uparrow$\ours  & Dynamic ($\delta_{kv}=0.95$) & \xmark  &  54.7\% & 3.4B & \textbf{51.50} \\
        $\uparrow$\ours  & Fixed ($r_{kv} = 512$) & \cmark  & 53.1\% & 3.4B & 50.89 \\
        $\uparrow$\ours  & Fixed ($r_{kv} = 512$) & \xmark  & 53.1\% & 3.4B & \textbf{51.63} \\
       
        \bottomrule
    \end{tabular}
    \caption{\small{Comparison of MLA with LayerNorm vs. without LayerNorm} }
    \label{tab:LN}
\end{table}

{Figure~\ref{fig:first loss plot} shows the loss curves of our SVD initialization with and without layer normalization, as well as with random initialization. The results demonstrate that removing layer normalization leads to lower loss values.}

\subsubsection{Larger Teacher or more Training Data?}
\label{ablation:data_vs_teacher}
Table~\ref{tab:data_vs_teacher} highlights the impact of increasing training data (tokens) versus using a larger teacher model on both accuracy score and training time. When using the same teacher model (Llama3.2-1B-Inst), increasing the number of training tokens (from 3.4B to 6.8B) improves performance but comes at the cost of significantly higher training time (from 4.82 to 9.64 hours).

On the other hand, switching to a larger teacher (e.g., Llama3.2-3B-Inst or Llama3.2-8B-Inst) provides notable accuracy improvements with less reliance on additional training data. For instance, using the 8B teacher with DPO achieves the highest score, outperforming training with double tokens under the smaller 1B teacher (55.13 vs. 53.42), even with less training time (8.96 vs. 10.06 hours). However, this comes with a moderate increase in training time.

These results suggest that leveraging a stronger teacher model is generally more efficient for improving accuracy than simply increasing training data. While additional tokens help, the benefit diminishes compared to the gains from using a larger teacher, making training with a larger teacher a more effective strategy when computational resources allow.

\begin{table*}[t]
     \setlength\extrarowheight{2pt}
    \centering
    \scalebox{1.0}{
    \begin{tabular}{llccc}
    \toprule
    Model and Setting & Teacher & Tokens & Training time & Avg Score \\
    \midrule
    \rowcolor{gray!15}
    \multicolumn{5}{c}{Base Model: \textbf{Llama3.2-1B-Inst};  100\% MLA Layers ($r_{kv}=512$, $r_q = 864$)}\\
    $\uparrow$\ours & Llama3.2-1B-Inst &  6.8B &  9.64 hours  &	52.12 \\ 
    $\uparrow$\ours + DPO & Llama3.2-1B-Inst &  7.0B &  10.06 hours &	53.42 \\ 
    $\uparrow$\ours & Llama3.2-1B-Inst &  3.4B & 4.82 hours & 51.63 \\ 
    $\uparrow$\ours + DPO & Llama3.2-1B-Inst &  3.6B &  5.24 hours  &	53.04 \\ 
    $\uparrow$\ours  & Llama3.2-3B-Inst &  3.4B &  6.24 hours & 52.25 \\ 
    $\uparrow$\ours + DPO  & Llama3.2-3B-Inst & 3.6B & 6.65 hours &		54.08 \\ 
    $\uparrow$\ours & Llama3.2-8B-Inst &  3.4B &  8.54 hours & 53.69\\ 
    $\uparrow$\ours + DPO & Llama3.2-8B-Inst &  3.6B &  8.96 hours &	55.13 \\
    \bottomrule
    \end{tabular}
    }
    \caption{\small{Comparison of training efficiency and accuracy when increasing training data (tokens) versus using a larger teacher model. Larger teachers yield better accuracy with moderate time increases. The time cost is measured on a 8 MI300 GPUs.}}
    \label{tab:data_vs_teacher}
\end{table*}

\end{document}